\documentclass[10pt,journal,compsoc]{IEEEtran}

\usepackage{algorithm}
\usepackage{algorithmic}

\usepackage{graphicx}
\usepackage{subfig}
\usepackage{booktabs}
\usepackage{array}
\usepackage{tabularx}
\usepackage{adjustbox}

\usepackage[table,dvipsnames]{xcolor}

\usepackage{amsmath,amsfonts,mathtools}
\usepackage{soul}
\usepackage{textcomp}

\usepackage{url}
\usepackage{cite}
\usepackage{hyperref} 

\usepackage{verbatim}
\usepackage{multicol}
\usepackage{multirow}
\usepackage{threeparttable}
\usepackage{stfloats}
\usepackage{xspace}
\usepackage{pgf-pie}
\usepackage{svg} 
\usepackage{listings}

\definecolor{nodecolor}{HTML}{de425b}     
\definecolor{attributecolor}{HTML}{9e64bd} 
\definecolor{valuecolor}{HTML}{31bd75}    
\definecolor{stringcolor}{HTML}{5580bd}   
\definecolor{commentcolor}{HTML}{5b44c2}  

\lstdefinelanguage{XML}{
  morekeywords={root,BehaviorTree,Sequence,TakeOff,SaveData,UseData,Land}, 
  morekeywords={[2]ID,saveto,usefrom,takeoff_target}, 
  keywordstyle=[1]\color{nodecolor}, 
  keywordstyle=[2]\color{black}, 
  alsoletter={:,-},                      
  morecomment=[s]{<?}{?>},               
  morecomment=[s]{<!--}{-->},            
  morecomment=[s]{<!}{>},                
  morestring=[b]",
  tag=[s]{<}{>},
  attributestyle=\color{green!60!black},
  attributevalue=\color{orange!80!black},
  stringstyle=\color{stringcolor},
  commentstyle=\color{gray}\itshape
}

\soulregister\cite7

\soulregister\eqref7

\newcommand{\sysname}{\textbf{\textit{SRDrone}}\xspace}

\begin{document}
\title{LLM-Driven Self-Refinement for Embodied Drone Task Planning}

\author{Deyu Zhang, \IEEEmembership{Member,~IEEE}, Xicheng Zhang, Jiahao Li,  Tingting Long, Xunhua Dai,  ~Yongjian Fu, Jinrui~Zhang,~\IEEEmembership{Member,~IEEE,}~Ju Ren, ~\IEEEmembership{Senior~Member,~IEEE},~Yaoxue~Zhang,~\IEEEmembership{Senior~Member,~IEEE}
	\IEEEcompsocitemizethanks{
		\IEEEcompsocthanksitem Deyu Zhang,  Xicheng Zhang, Jiahao Li, Tingting Long, Xunhua Dai, Yongjian Fu  are with School of Computer Science and Engineering, Central South University. Changsha 410083, China. E-mails:  \{zdy876, csu\_xichengzhang, jiahao\_li, tingtinglong, dai.xh,  fuyongjian\}@csu.edu.cn
		\IEEEcompsocthanksitem Jinrui Zhang,  Ju Ren, Yaoxue Zhang are with the Department of Computer Science and Technology, Tsinghua University, Beijing 100084, China. E-mails: \{jinruizhang, renju, zyx\}@tsinghua.edu.cn
		\IEEEcompsocthanksitem Corresponding author: Jinrui Zhang 
	}

}
\markboth{IEEE Transactions on Mobile Computing,~Vol.~XX, No.~XX, XXX~2021}
{Deyu Zhang \MakeLowercase{\textit{et al.}}: Real-Time Multi-Person Pose Estimation based with Motion Vector on Mobile Devices}

\IEEEtitleabstractindextext{%
\begin{abstract}
We introduce \textit{\sysname}, a novel system designed for self-refinement task planning in industrial-grade embodied drones. \textit{\sysname} incorporates two key technical contributions:
First, it employs a continuous state evaluation methodology to robustly and accurately determine task outcomes and provide explanatory feedback. This approach supersedes conventional reliance on single-frame final-state assessment for continuous, dynamic drone operations.
Second, \textit{\sysname} implements a hierarchical Behavior Tree (BT) modification model. This model integrates multi-level BT plan analysis with a constrained strategy space to enable structured reflective learning from experience.
Experimental results demonstrate that \textit{\sysname} achieves a 44.87\% improvement in Success Rate (SR) over baseline methods.
Furthermore, real-world deployment utilizing an experience base optimized through iterative self-refinement attains a 96.25\% SR. By embedding adaptive task refinement capabilities within an industrial-grade BT planning framework, \textit{\sysname} effectively integrates the general reasoning intelligence of Large Language Models (LLMs) with the stringent physical execution constraints inherent to embodied drones. \textit{Code is available at \href{https://github.com/ZXiiiC/SRDrone}{\texttt{https://github.com/ZXiiiC/SRDrone}}.}

\end{abstract}

\begin{IEEEkeywords}
Edge Intelligence, Embodied Drones, Task Planning, Behavior Tree
\end{IEEEkeywords}}

\maketitle

\IEEEdisplaynontitleabstractindextext
\IEEEpeerreviewmaketitle

\section{Introduction}
As the pivotal platform for the low-altitude economy, drones are fundamentally shaping the scale and effectiveness of this emerging industry\cite{huang2025small}.  In this context, intelligent task planning is critical for drone deployment, demanding core capabilities such as autonomous decision-making, complex task comprehension, dynamic environment adaptability, and natural human-machine interaction. Key application scenarios, e.g., smart grid/pipeline inspection\cite{tang2024industry}, urban logistics delivery\cite{serrano2021selecting}, and emergency response\cite{agrawal2020next}, further necessitate reliable autonomous execution of complex missions. Recent breakthroughs in Large Language Models (LLMs) and multimodal extensions significantly enhance the cognitive intelligence potential of drone control systems by leveraging vast knowledge bases and robust semantic reasoning capabilities~\cite{chen2023typefly,piggott2023net,wang2025multi}.

Despite deploying LLMs for automated planning leveraging their powerful reasoning capabilities to generate complex plan, several critical limitations remain in practical applications. (1) \textbf{High Dependence of Human Expertise.} Other methods leverage LLMs to generate an initial drone path plan.  To achieve adaptability in dynamic operational environments, these approaches rely critically on human experts to perform dynamic online adjustments during mission execution~\cite{ao2024llm,chen2023typefly,parakh2024lifelong}. Yet, this dependence on continuous manual intervention entails prohibitively high operational costs.  Furthermore, expert-driven adjustments inherently compromise critical performance attributes, including dynamic responsiveness, persistent task execution, and terrain adaptation. This degradation arises from fundamental limitations associated with human factors: cognitive processing delays, finite operator bandwidth leading to saturation, unstable communication links, and the rigidity inherent in transferring expertise to dynamic scenarios. (2) \textbf{Inadequate Generalization and Dynamic Adaptability.}
The predominant approaches employ LLMs to statically generate Behavior Trees (BTs), exploiting their industrial compatibility to produce executable specifications intended for one-time deployment~\cite{izzo2024btgenbot,chen2024integrating}. However, such static BTs fundamentally limited adaptability to Out-of-Distribution (OOD) environments. Consequently, when deployed in unfamiliar settings or subjected to unforeseen environmental conditions (e.g., unexpected weather), the performance of these static BTs exhibits significant degradation, compromising the reliability of execution in practical scenarios.


To address these limitations, we propose \sysname, a framework designed to enable autonomous and scalable drone task planning without relying on human experts. \sysname employs continuous motion state analysis for autonomous performance evaluation of mission execution, thereby activating a self-reflective iteration mechanism that removes the requirement for real-time human supervision. Furthermore, it utilizes hierarchical behavior tree modification with semantic understanding of behavioral structures. This enables online dynamic correction of LLM-generated BTs for robust adaptation in OOD environments.


However, designing the framework faces two key challenges: \textbf{1) How to achieving reliable self-assessment solely based on execution feedback, without human oversight?} It requires the LLMs to interpret complex, multi-modal state information, accurately detect deviations from the intended goal, and provide trustworthy evaluations supported by explanatory reflections.Achieving the automation, reliability, and robustness of this self-assessment loop constitutes a fundamental barrier to realizing genuine task autonomy. 


\textbf{2) How to convert unstructured textual feedback from the LLMs into precise and structured modifications for the BTs?} Establishing a stable mechanism for iterative BT refinement necessitates ensuring logical consistency and verifying operational feasibility within this transformation. This involves reconciling the substantial semantic gap between informal language descriptions and the formal syntax and semantics of tree-based structures.

To tackle the above challenges, we integrate several new techniques in \sysname, each addressing one of the above challenges. 
1) \textbf{Continuous State Evaluation}. Onboard drone sensors capture extensive datasets reflecting mission execution. However,  these raw sensor streams are typically voluminous and lack explicit semantic interpretability concerning the underlying drone actions. We posit that extracting the latent action semantics embedded within this data is crucial for effective self-assessment. To achieve this, we propose the Continuous Motion and Spatial Reasoning (CMSR) algorithm. CMSR explicitly reasons about the drone's ego-motion trajectory and its spatiotemporal relationships with the surrounding environment. Moreover, the algorithm employs a selective focus mechanism to analyze only critical states within the data stream, significantly mitigating the computational overhead associated with continuous reasoning.
2) \textbf{Hierarchical BT Modification}. While LLMs generate improvement plans through reflection experience, a significant gap exists between the loosely unstructured textual experience and the syntactically rigorous BTs. Our core insight is that bridging this gap requires structured and precise analysis of plan failures to iteratively refine the BTs. To achieve this, we design \textit{Hierarchical BT Modification} to perform  multi-level semantic analysis of the current BT plan to isolate failure points. It then synthesizes precise node-level modification suggestions by leveraging a constrained strategy space. This transforms abstract LLM reflections into executable, fine-grade refinement actions.

We conduct comprehensive experiments across four scenarios: path planning, object searching, obstacle navigation, and complex composite tasks. In software-based and hardware-in-the-loop simulations, \sysname achieved a 44.87\% higher average Success Rate (SR) compared to state-of-the-art baselines. Subsequent real-world deployment on physical drones validated its operational efficacy, attaining a 96.25\% task success rate. This high performance is enabled by the system's capacity for iterative self-refinement of its experience base, initially derived from simulation data. The results demonstrate effective simulation-to-reality transferability, concurrently mitigating the risks and costs associated with drone damage due to real-world mission failures. When evaluating the accuracy of plan-level failure explanations, our continuous state evaluation method achieved 80.07\% accuracy. This significantly outperformed conventional final-state-checkpoint methods, which yielded only 12.18\% accuracy.


In summary, our work makes the following main contributions:
\begin{enumerate}
    \item We propose \sysname, to the best of our knowledge,  the first framework that enables self-evolving BTs for drone task planning. Our approach overcomes adaptability limitations in LLM-based drone systems via a closed-loop refinement cycle that facilitates continuous field adaptation without human intervention.
    \item We introduce a time-series assessment technique for self-supervised failure diagnosis, and propose a  hierarchical BTs repair mechanism that translates unstructured LLM feedback into formal syntax modifications for structural evolution.
    \item We implement and evaluate \sysname through both simulated experiments and real-world deployment, demonstrating its effectiveness via comparisons with SOTA methods across four representative scenarios and validating its practical feasibility. 
\end{enumerate}
\section{Background and Motivation}
\subsection{Behavior Tree}
Behavior Trees constitute a robust, hierarchical decision-making architecture prevalent in robotics and AI\cite{ghzouli2023behavior,ruifeng2019research,chen2024integrating}, fundamentally structuring agent behavior as a directed tree of interconnected nodes—primarily Control Nodes (e.g., Sequence: execute children sequentially until failure; Fallback/Selector: execute children sequentially until success; Parallel: execute children concurrently) that govern execution flow, Execution Nodes (e.g., Action nodes: perform state-changing tasks like movement commands; Condition nodes: evaluate true/false predicates about the system or environment) that define atomic behaviors and checks, Decorators to modify child node properties (e.g., looping, inverting result), and reusable Subtrees—all orchestrated through repeated "ticking" from the root node, propagating ticks based on node logic and returning statuses (Success, Failure, Running) to dynamically select actions in response to real-time changes in a shared data repository (the Blackboard). 

 In drone  flight control, BTs provide the high-level decision logic, decomposing complex missions into manageable tasks\cite{ghzouli2020behavior,ogren2012increasing}; for instance, an Action node might command flight maneuvers like "move to waypoint", while concurrent Condition nodes monitor critical factors such as battery level or obstacle proximity, allowing the BT to dynamically adjust the drone's actions (e.g., triggering a return-to-home procedure if battery is low) by succeeding or failing specific branches based on sensor feedback.

\subsection{Self-Reflection Capability of LLM}
LLM self-reflection denotes the capability to critically evaluate their own outputs based on external feedback and iteratively optimize subsequent results\cite{shinn2023reflexion,liu2023reflect}. This reflective process requires two inputs: the model's prior generated content or action sequences requiring assessment, such as text passages, code snippets, or decision sequences; and externally provided feedback, including task outcomes, explicit error signals, or explicit critiques and guidance\cite{wadhwa-etal-2024-learning-refine,chen-etal-2024-prompt}. The output comprises strategically generated textual content facilitating improvement, such as root-cause analyses of errors, actionable optimization proposals, and ultimately yielding revised or enhanced generated content or behaviors through the application of these refined strategies.

Compared to traditional paradigms like Reinforcement Learning (RL)\cite{peiyuan2024agile,rita-etal-2024-countering,ahn2022can} and Imitation Learning (IL)\cite{sun2024prompt}, LLM self-reflection offers significant advantages. RL relies on meticulously designed scalar reward functions, which often provide sparse, imprecise signals regarding complex behavioral defects, leading to low sample efficiency and demanding vast interaction data for fine-tuning. IL necessitates costly, high-quality, comprehensive expert demonstration data, which constrains generalization. In contrast, LLM self-reflection leverages natural language feedback—a format inherently richer and more expressive than scalar rewards, conveying substantially higher information density\cite{sumers2021learning,wu2023fine}. Furthermore, the process is driven by the LLM's internal linguistic reasoning, typically enabling rapid behavioral adjustment during inference through iterative generation and evaluation without requiring computationally expensive weight updates or retraining, thus offering enhanced flexibility and computational efficiency\cite{shinn2023reflexion}.

\subsection{A Motivation Study}
\label{sec:motivation}
The self-reflection capabilities of LLMs open new opportunities for enabling truly autonomous and scalable drone task planning. However, during our preliminary experiments, we find two critical issues remain unresolved: (1) LLMs' tendency to misjudge task outcomes without human feedback; and (2) the low error correction rates resulting from unstructured textual reflection experiences.

\subsubsection{Necessity of Continuous State Evaluation for Plan-level Failure Detection and Explanation}

\begin{figure}[t]
    \centering
    \includegraphics[width=0.5\textwidth]{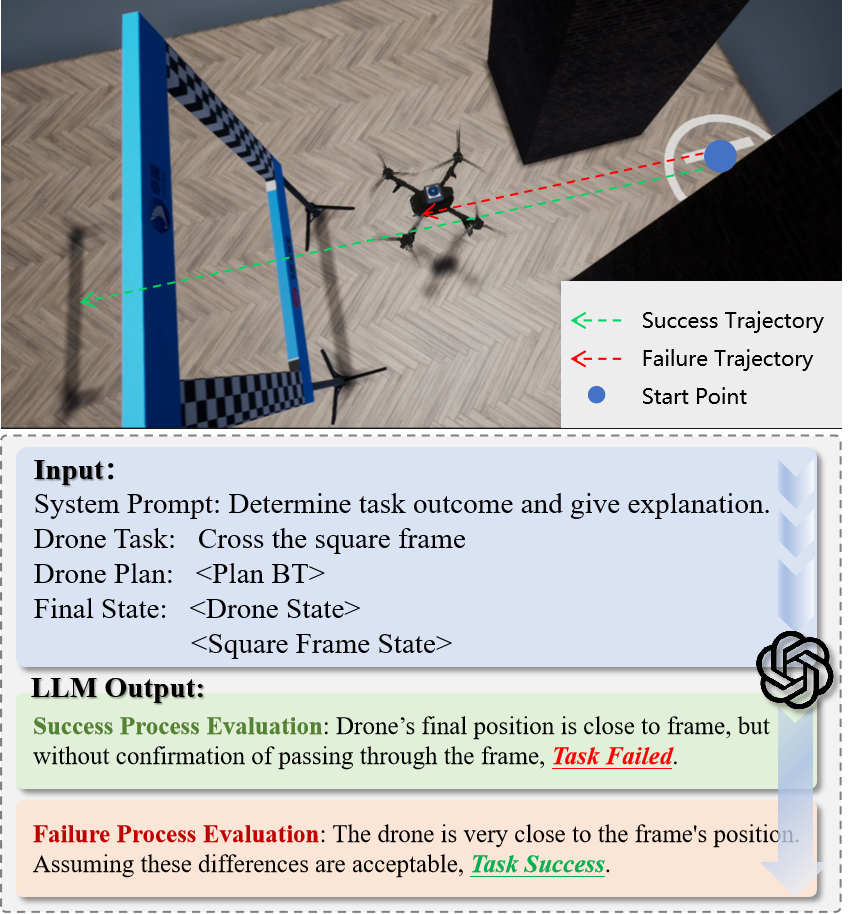} 
    \caption{Evaluation results from Final-State-Only method on successful and failed trajectories in cross square frame task. \textcolor{CornflowerBlue}{\textbf{Blue}}: Input prompt. \textcolor{OliveGreen}{\textbf{Green}}: Successful trajectory misjudged as failures. \textcolor{Red}{\textbf{Red}}: Failed trajectory misjudged as successful.}
    \label{fig:motivation1} 
\end{figure}

Unlike human-involved approaches where experts monitor the entire execution process, existing self-supervised failure detection methods commonly rely on single-frame final state to assess plan-level embodied intelligence task outcomes\cite{liu2023reflect,das2021semantic}. However, this paradigm proves unsuitable for plan-level drone missions failure evaluation. Plan-level failures stem from fundamental flaws in the plan's design, such that even perfect execution cannot ensure mission success. As shown in Figure~\ref{fig:motivation1}, final-state-only evaluation induces bidirectional diagnostic errors: it systematically misclassifies failed trajectories as successful while misclassifying correct trajectories as failed.

In drone operations, reliance on single-frame terminal states for mission evaluation induces erroneous outcome judgments and misleading explanations. This limitation arises because terminal states merely capture static goal achievement, neglecting the continuous spatiotemporal fidelity inherent in trajectory execution, which demands integrated temporal assessment. Consequently, a temporal-spatial integrated evaluation framework becomes essential for accurate outcome determination and interpretable failure analysis. This framework decouples analytical objectives: (1) Trajectory compliance verification (assessing adherence to planned motion paths).
(2) Geo-referenced positioning monitoring (evaluating environment-relative positioning). This dual approach significantly enhances evaluation reliability while ensuring cross-scenario robustness.

Thus, in Section~\ref{sec:evaluator}, we devise a continuous state evaluation method that processes onboard sensor streams to perform semantic analysis of task-execution processes, enabling robust outcome determination and interpretable failure diagnostics.

\subsubsection{Necessity of Structured Error Correction in Behavior Tree Self-Reflection}

Existing self-reflection frameworks generate unstructured textual experiences that are incompatible with the formal syntax of Behavior Trees, resulting in low correction rate for BT plan flaws. As shown in Table~\ref{tab:bt_errors}, coarse-grained reflection mechanisms consistently fail to address hierarchical dependencies during BT repair: our experiments reveal only 39.28\% success in detecting missing dependencies and 28.6\% in correcting invalid control flows across 28 repair attempts. These results highlight a critical limitation: unstructured textual outputs inherently lack architectural constraint encoding for Behavior Trees. Consequently, current approaches merely patch isolated node actions while neglecting inter-node dependencies. Unstructured reflection formats are fundamentally incapable of correcting architectural flaws due to their inherent inability to encode hierarchical dependencies. For instance, unstructured text fails to formally express missing or invalid node connections-fundamental requirements for valid BTs execution. 

\begin{table*}[t]
\centering
\small
\renewcommand{\arraystretch}{1}
\setlength{\tabcolsep}{7pt} 
\arrayrulecolor{black!70} 
\caption{Correction rate of existing reflection methods for two typical behavior tree errors}
\begin{adjustbox}{center}
\begin{tabular}{>{\color{black!80}}l!{\vrule height 0.8\normalbaselineskip depth 0pt width 0.4pt} >{\color{black!80}}l!{\vrule height 0.8\normalbaselineskip depth 0pt width 0.4pt} >{\raggedright\arraybackslash}p{6cm}!{\vrule height 0.8\normalbaselineskip depth 0pt width 0.4pt} >{\centering\arraybackslash}c}
\toprule[0.8pt]
\multicolumn{1}{c|}{\textbf{Error Type}} & \multicolumn{1}{c|}{\textbf{BT Manifestation}} & \multicolumn{1}{c|}{\textbf{Example}} & \multicolumn{1}{c}{\textbf{Correction Rate}} \\
\midrule[0.5pt]
Missing Dependency 
  & Absence of precondition nodes 
  & Lack actions for searching objectives 
  & 39.28\% \\[\dimexpr0.5em-\arrayrulewidth]
Invalid Control Flow 
  & Mismatched composite node logic 
  & Using \textit{Sequence} where \textit{Fallback} required 
  & 28.6\% \\
\bottomrule[0.8pt]
\end{tabular}
\end{adjustbox}
\normalsize
\label{tab:bt_errors}
\end{table*}

To address this gap, in Section \ref{sec:bt_correction}, we propose Hierarchical BT modification, a structured correction framework comprising two sequential stages. Firstly, hierarchy BT plan analysis systematically identifies structural flaws by analyzing execution traces and formalizing them as repair constraints. Secondly, node-level precise modification applies targeted corrections to individual nodes while ensuring compositional integrity. This fine-grained modification approach enables atomic action correction for primitive node flaws and architectural constraint enforcement to preserve the hierarchical structure of the BT.

\section{SSDrone}

\subsection{Overview}
\sysname aims to achieve two key goals: (1) robustly evaluate task execution outcomes to detect planning flaws, and (2) drive progressive refinement to ensure successful task completion. Figure~\ref{framework} illustrates the overall architecture of \sysname, which consists of two main phases:

\begin{figure*}[t] 
    \centering
    \includegraphics[width=\textwidth]{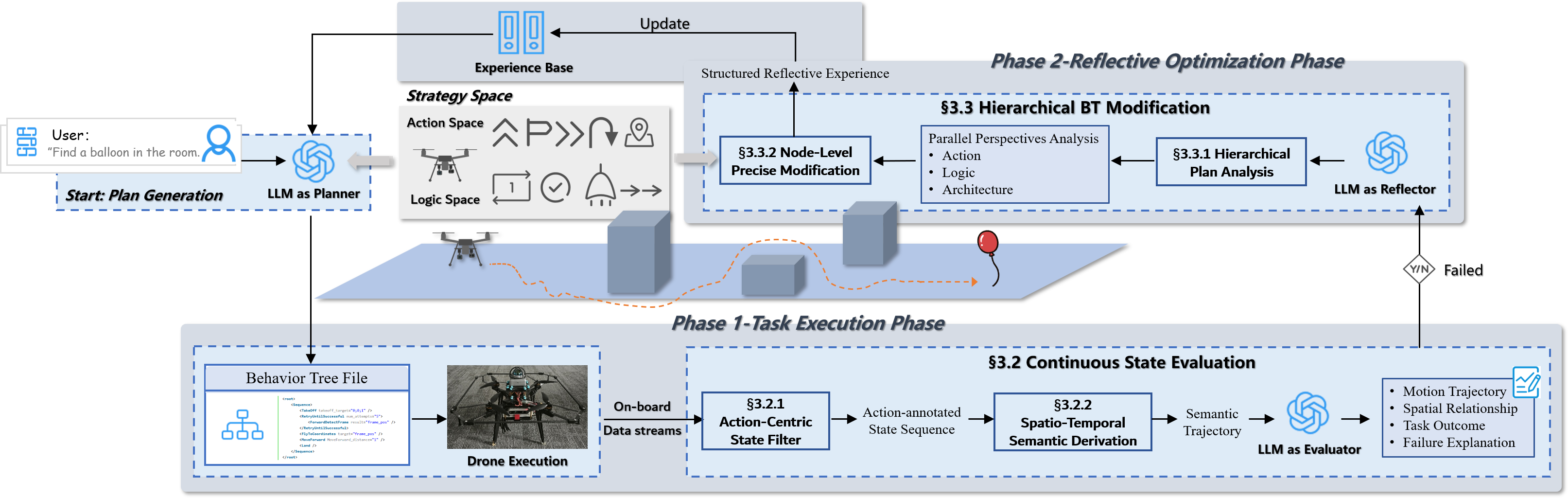} 
    \caption{The overall framework of \sysname.}
    \label{framework}
\end{figure*}

(1) During \textit{Task Execution Phase}, we design a Continuous State Evaluation framework which enables reliable outcome determination and interpretable anomaly attribution for drone operations. This framework initiates with an Action-Centric State Capture module that filters critical flight states from high-frequency drone data streams. Building upon this filtered data, our proposed CMSR algorithm performs spatiotemporal semantic extraction, converting multidimensional temporal sensor data into natural language task narratives processable by LLMs. The framework culminates in task determination and failure explanation, generating interpretable diagnostic insights to support subsequent planning optimization.

(2) In the \textit{Reflective Optimization Phase}, \sysname identifies and resolves planning flaws using a Hierarchical BT Modification approach. This method facilitates a two-stage refinement process to generate structural reflective experience and guarantee the reliability of refinements. The first stage conducts hierarchical analysis of the existing BTs to localize errors across behavioral execution, logical conditions, and planning structure. Subsequently, error analysis results are transformed into actionable modifications through dual-constraint processing: operational feasibility within hardware/software boundaries and structural validity adhering to BT architecture standards. This approach generates precise node-level correction specifications for behavioral and logical nodes. Each iteration cycle updates the Experience Base with validated corrections, while subsequent planning iterations leverage this refined knowledge base to optimize the BTs, establishing a closed-loop improvement mechanism for continuous planning enhancement. 

\subsection{Continuous State Evaluation}
\label{sec:evaluator}

The reliability of autonomous drone task refinement depends critically on robust failure detection and interpretable error diagnosis. Existing approaches relying on single-frame final-state assessment\cite{liu2023reflect,das2021semantic} prove fundamentally inadequate for continuous drone operations (Section\ref{sec:motivation}). This limitation manifests when tasks yield geometrically similar terminal states but exhibit diametrically opposed execution processes . Such critical behavioral distinctions remain undetectable through snapshot-based evaluation.

To enable continuous execution trajectory assessment, we need to address two key problems:
(i) \textbf{High-frequency Data Streams} Drone control systems demand $>$30~Hz actuation signals, generating high-velocity sensor data streams that rapidly exceed the context window capacity of state-of-the-art Large Language Models.(ii) \textbf{Temporal Data Semantics Extraction} LLMs exhibit inherent limitations in processing continuous numerical time-series data due to their discrete token-based architecture\cite{chang2025llm4ts}\cite{jin2023time}. This architectural gap impedes effective modeling of long-range temporal dependencies and complex dynamic patterns essential for task-critical diagnostics\cite{zeng2023transformers}.

\subsubsection{Action-Centric State Filtering}
\label{sec:action_state}
\begin{figure}
    \centering
    \includegraphics[width=\linewidth]{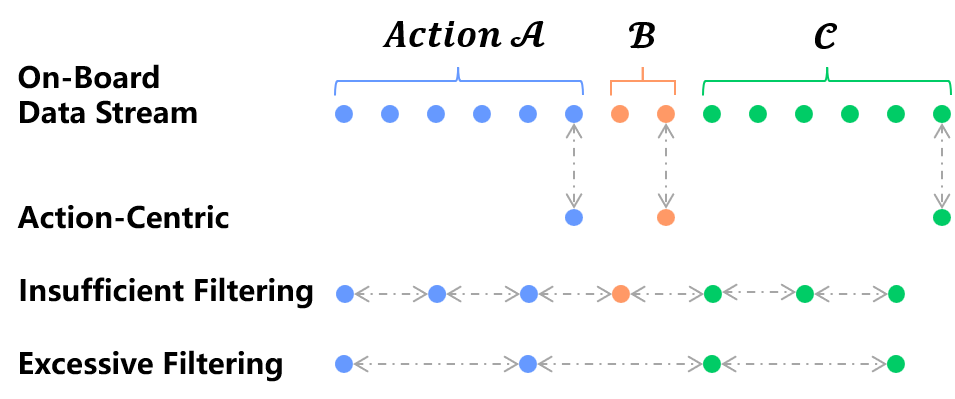}
    \caption{Comparison of different data filtering methods: Fixed-interval filtering results in insufficient filtering, which generates redundant data, or excessive filtering, which risks losing critical states. Our Action-Centric Filtering adaptively balances semantic intent preservation with computational efficiency, avoiding rigid interval constraints.}
    \label{fig:filter}
\end{figure}

Autonomous task evaluation requires state capture at granularities that preserve semantic intent while respecting computational constraints.Drone sensors produce high-frequency data streams (typically $>$30~Hz) that require filtering for downstream processing. For instance, motor control signals in platforms like Pixhawk\cite{pixhawk} operate at 100~Hz, while IMU sensors commonly generate data at 50~Hz. However, as shown in Figure~\ref{fig:filter}, fixed-interval sampling faces two limitations: 1) \textbf{Excessive filtering} can result in the omission of critical states due to large gaps between recorded samples. 2) \textbf{Insufficient filtering} introduces redundant data without added semantic value, as raw sensor readings lack explicit behavioral context. For instance, a 1$Hz$ sample might skip the \texttt{FlyToCoordinates} $\rightarrow$ \texttt{Land} transition entirely, while 20$Hz$ sampling captures repetitive pose updates without distinguishing action semantics. This dual challenge, involving the loss of behavioral fidelity at low sampling rates and the presence of semantic ambiguity at high sampling rates, underscores the need for action-aligned state recording to maintain both temporal resolution and behavioral intent.

To resolve this problem, we introduce \textit{action-triggered state recording}-capturing execution states $s_i$ exclusively at action completion boundaries. Action-triggered recording leverages the inherent characteristics of drone control systems. Controllers execute commands as discrete semantic units (e.g., \texttt{Takeoff}, \texttt{FlyToCoordinates}), with state transitions concentrated at action completion boundaries. These boundaries naturally align with kinematically significant configuration changes where positional and orientational shifts manifest most distinctly. By eliminating intra-action sampling, this approach achieves data reduction while preserving all critical state transitions essential for task verification. The resulting triple alignment-semantic, kinematic, and computational-ensures captured states reflect meaningful behavioral endpoints rather than arbitrary motion intermediates.

The process works as follows:
(1) Temporal Anchoring: Record state vectors $s_i = (x,y,z,\psi)$ exclusively at the completion of each action node $a_i$, where pose $(x,y,z)$ and yaw $\psi$, follow NWU convention.
(2) Failure-driven Logging: For failed actions, augment $s_i$ with failure tag to enable error tracing.
(3) Modality-aware Capture:

\begin{equation}
\mathcal{D}_i = 
\begin{cases}
s_i & \text{(motion control nodes)} \\
s_i \oplus \mathcal{E}_i & \text{(information capture nodes)}
\end{cases}
\end{equation}
where $\mathcal{E}_i$ denotes environmental observations from the drone's onboard computing platform. Motion control nodes (e.g., \texttt{MoveForward}, \texttt{FlyToCoordinates}) alter the drone's own pose, while information capture nodes (e.g., \texttt{ForwardDetectBalloon}) acquire environmental information.
The output $\Pi = (\pi_1, \pi_2, \dots, \pi_n)$ forms an action-annotated state sequence, where each element $\pi_i = (a_i, \mathcal{D}_i)$ combines an action $a_i$ with its corresponding state-data pair $\mathcal{D}_i$.
\subsubsection{Spatio-Temporal Semantic Derivation}
\label{sec:semantic_extraction}

Building upon the Action-Centric State Capture (Section \ref{sec:action_state}), a core challenge in drone task evaluation lies in analyzing the execution process, specifically discerning both the drone's intrinsic motion behaviors and its spatio-temporal relationships with the environment. However, directly utilizing raw trajectory data for LLM-based analysis faces a fundamental barrier: LLMs' inherent limitations in comprehending continuous numerical time-series data due to their discrete token-based architecture, as established in prior work \cite{chang2025llm4ts}\cite{jin2023time}. To tackle the prevailing limitation, our Continuous Motion and Spatial Reasoning (CMSR) algorithm processes action-annotated data through concurrent semantic derivation streams that extract ego-motion behavior patterns while simultaneously inferring spatial environmental interactions, with the motion output directly informing spatial reasoning.

\textbf{Ego-Motion Semantic Derivation.} CMSR initiates by extracting the drone's intrinsic motion characteristics from sequential state frames. For each consecutive state pair $(\pi_t, \pi_{t+1})$, the algorithm derives the motion semantics through spatial-temporal analysis:
\begin{equation}
\vec{\mathcal{M}}_t = \text{DeriveMotion}(\pi_t, \pi_{t+1})
\end{equation}
where $\vec{\mathcal{M}}_t$ represents a semantic motion vector:
\begin{equation}
\vec{\mathcal{M}}_t = \begin{bmatrix}
\text{behavior\_type}\\
\text{displacement} \\ 
\text{orientation\_change} \\ 
\end{bmatrix}
\end{equation}

The DeriveMotion process is explicitly designed to translate low-level trajectory data into high-level, LLM-compatible semantics. To overcome LLMs' inherent difficulties in interpreting continuous numerical time-series, it encodes raw motion into structured symbolic representations:
\begin{itemize}
\item \text{behavior\_type}: Classifies the discrete action type to provide immediate contextual intent, bypassing complex motion pattern recognition.
\item \text{displacement}: Translates relative position changes into human-readable directional descriptions, replacing dense coordinate sequences with concise spatial relationships.
\item \text{orientation\_change}: Encodes yaw variation as rotational semantics, abstracting continuous angular data into discrete, meaningful directional shifts.
\end{itemize}

These semantic motion primitives are temporally integrated to form the ego-trajectory narrative:
\begin{equation}
\mathcal{T}_{\text{ego}} = \bigoplus_{t=0}^{T-1} \vec{\mathcal{M}}_t
\end{equation}
This reconstruction preserves continuous motion semantics essential for understanding execution behaviors, translating raw coordinates into textual descriptions:
\begin{quote}
\textit{"During $[t_2, t_3]$:
FlytoCoordinates, the drone fly to coordinate NWU(1, 2, 1.5)"\\
"During $[t_3, t_4]$:
TurnLeft, the drone turn left with 90\textdegree clockwise rotation"\\
"During $[t_4, t_5]$:
MoveForward, the drone move forward 1m to NWU(1, 3, 1.5)"
}
\end{quote}

\textbf{Environmental Relationship Reasoning.} The spatial relationships between drones and their environments play a critical role in determining mission outcomes. These relationships encompass essential safety constraints and operational requirements, forming the foundation for evaluating task success. However, current LLMs’ abilities in spatial reasoning remain relatively unexplored.\cite{wu2024mind}. Their inherent limitations in geometric reasoning and fine-grained 3D computations hinder the derivation of meaningful spatial semantics from coordinate transformations. This bottleneck prevents direct application of standard LLM capabilities to tasks requiring spatial analysis.

 
\begin{algorithm}[H]
\caption{Continuous Motion and Spatial Reasoning (CMSR)}
\label{alg:cmsr_semantics}
\begin{algorithmic}[1]
\REQUIRE Action-annotated state sequence $\Pi = \{\pi_1,...,\pi_T\}$ 
\ENSURE Semantic trajectory $\mathcal{T}_{\text{sem}}$
\STATE $\mathcal{T}_{\text{ego}} \gets \emptyset$; $\mathcal{R} \gets \emptyset$; $\mathcal{O} \gets \emptyset$  
\FOR{$t \gets 2$ to $T$}  
    \STATE $\vec{\mathcal{M}}_t \gets \text{DeriveMotion}(\pi_{t-1}, \pi_{t})$ \hfill \textit{// Ego-motion semantic derivation between consecutive states}
    \STATE $\mathcal{T}_{\text{ego}} \gets \mathcal{T}_{\text{ego}} \oplus \vec{\mathcal{M}}_t$ \hfill \textit{// Temporal integration}
    
    \STATE $\mathcal{O} \gets \text{UpdateEnvironmentState}(\mathcal{O}, \pi_t)$ \hfill \textit{// Update environment state with current observation}
    
    \STATE $\mathcal{R}_t \gets \emptyset$
    \FORALL{$o \in \mathcal{O}$}  
        \STATE $r_t^o \gets \text{InferSpatialRelation}(\vec{\mathcal{M}}_t, o)$ \hfill \textit{// Spatial relation reasoning with environment object}
        \STATE $\mathcal{R}_t \gets \mathcal{R}_t \cup \{(o, r_t^o)\}$
    \ENDFOR
    \STATE $\mathcal{R} \gets \mathcal{R} \cup \{(t, \mathcal{R}_t)\}$ \hfill \textit{// Store temporal-spatial relationships}
\ENDFOR
\STATE $\mathcal{T}_{\text{sem}} \gets \text{Align}(\mathcal{T}_{\text{ego}}, \mathcal{R})$ \hfill \textit{// Spatio-temporal fusion}
\RETURN $\mathcal{T}_{\text{sem}}$
\end{algorithmic}
\end{algorithm}

To this end, we propose a heuristic framework for spatial reasoning between drones and environments. We firstly instruct the LLM to maintain a persistent environment state representation $\mathcal{O}$ through heuristic chain-of-thought reasoning. At each time step $t$, the LLM integrates the current observation $\mathcal{E}_t$ with temporal context using:
\begin{equation}
\mathcal{O} \gets \text{UpdateEnvironmentState}(\mathcal{O}, \pi_t)
\end{equation}

This cognitive integration process enables $\mathcal{O}$ to adaptively track objects with evolving states while incorporating entities that re-enter perceptual blind spots, thereby maintaining holistic environment awareness through spatiotemporal reasoning.

The updated environment state $\mathcal{O}$ then enables object-centric spatial reasoning by establishing relations between ego-motion semantics $\mathcal{M}_t$ and individual environmental entitie $o\in \mathcal{O}$:
\begin{equation}
r_t^o \gets \text{InferSpatialRelation}(\vec{\mathcal{M}}_t, o)
\end{equation}

The InferSpatialRelation module employs a context-aware approach using LLM to reason about the drone's spatial relationships based on environmental information. It exploits the inherent characteristics of spatial relationship semantics by modeling the intrinsic properties of drone-environment interactions, generating tuple outputs $r_t^o = (\phi_{\text{intent}}, \phi_{\text{proximity}}, \phi_{\text{safety}})$ where each $\phi$ represents a semantic descriptor. These outputs enable precise contextual reasoning for task evaluation through three key operations:
\textbf{1) Navigational Intent}: Determines target approach vectors (e.g., "aligning with landing marker") by comparing drone motion trends with target coordinates, translating trajectory patterns into intentional navigation semantics for task completion assessment.
\textbf{2) Proximity Awareness}: Quantifies spatial relationships with key waypoints (e.g., "within 2m of checkpoint X") by analyzing distances to critical coordinates, transforming positional data into actionable proximity metrics for execution precision evaluation.
\textbf{3) Collision-Centric Safety}: Detects obstacle avoidance states (e.g., "circumnavigating obstacle B") through geometric containment checks considering drone dimensions, generating safety semantics primarily for virtual environment trajectory adjustment.

The final semantic trajectory $\mathcal{T}_{\text{sem}}$ integrates both egocentric motion ($\mathcal{T}_{\text{ego}}$) and environmental spatial relationships ($\mathcal{R} = \{(t, \mathcal{R}_t)\}$ where $\mathcal{R}_t = \{(o, r_t^o)\}$) into LLM-compatible narratives:
\begin{quote}
\textit{"During $[t_8, t_9]$: The drone MoveForawrd 1m to NWU(-0.93, 1.31, 1.26) while traversing square frame at NWU(-0.86, 1.29, 1.18)."}
\end{quote}

This spatio-temporal semantic derivation provides the foundational representation for continuous state evaluation, enabling LLM-based interpretation of integrated motion and environmental relationship patterns.

\subsection{Hierarchical BT Modification}
\label{sec:bt_correction}

\sysname’s ability to autonomously adapt hinges on dynamically modifying its LLM-generated Behavior Tree. In case of uncorrected static BTs operating in Out-of-Distribution (OOD) environments, the pre-defined BTs become misaligned with the unpredictable environment, leading to significant degradation in mission performance and loss of autonomy.

Existing approaches rely on human experts to provide precise adjustments to the Behavior Tree (BT) during execution\cite{ao2024llm}. However, this manual intervention contradicts our design goal of full autonomy. While leveraging LLMs for self-reflective BT correction presents a opportunity to eliminate human dependency, our motivation study (Section \ref{sec:motivation}) reveals a critical limitation: prevailing LLM-based self-reflection methods, only use free-form textual outputs, are unsuited to the rigorous syntactic constraints and complex hierarchical structure inherent to Behavior Trees

To enable BTs adaptation, this section introduces \sysname’s hierarchical semantic modification framework for correcting LLM-generated Behavior Trees in constrained strategy space.

\subsubsection{Hierarchical Plan Analysis}
\begin{figure}[!t]
    \centering
    \includegraphics[width=\columnwidth]{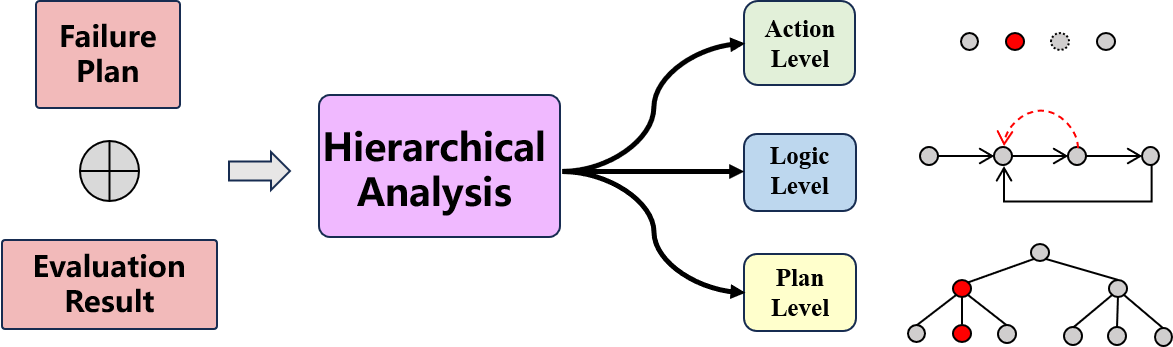}
    \caption{The illustration of hierarchical plan analysis.}
    \label{fig:hierarchy_analysis}
\end{figure}

Based on evaluation results and the current failure plan, the framework establishes a hierarchical analysis mechanism that diagnoses BT deficiencies from three parallel perspectives, as illustrated in Figure~\ref{fig:hierarchy_analysis}:
\begin{itemize}
    \item \textit{Action Layer} ($\alpha$): At the individual node level, verifies whether action nodes are correctly selected and identifies missing critical atomic actions in leaf nodes.
    \item \textit{Logic Layer} ($\lambda$): At the node relationship level, assesses if action nodes are organized with appropriate control logic and detects improper logical control flows in intermediate nodes.
    \item \textit{Mission Layer} ($\Pi$): At the plan-task level, analyzes potential macro-level misunderstandings between the composite plan and task requirements.
\end{itemize}
This hierarchical analysis aligns with the inherent architecture of Behavior Trees by decomposing their complex structure into semantically distinct layers while preserving the original design principles. The process creates a bidirectional mapping between the hierarchical structure and mission objectives, enabling both fine-grained diagnosis at execution level and high-level understanding of planning intent. By maintaining this structural-semantic correspondence, the analysis not only provides precise localization of misalignment sources but also establishes a contextual reference for subsequent modifications, ensuring continuity with the original plan's intent.

\subsubsection{Node-level Precise Modification}
As demonstrated in Section \ref{sec:motivation}, coarse-grained reflective experience exhibit limited effectiveness in rectifying structural defects of BT plans. To enhance the executability and efficiency of our reflection mechanism, this section proposes a strategy-space constrained correction framework with structured output formulation.

we firstly define a composite strategy space comprising \textit{Action Space} ($\mathcal{A}$) and \textit{Logic Space} ($\mathcal{L}$). Specifically, the Action Space is determined by drone-specific hardware configurations - different sensor suites (e.g., LiDAR vs. RGB cameras) and computational platforms (e.g., embedded GPUs vs. microcontrollers) yield distinct $\mathcal{A}$ spaces, as they impose physical constraints on feasible actions. Meanwhile, the Logic Space is defined by BT syntax specifications, encompassing supported control flow primitives such as sequence nodes, selector nodes, and decorator patterns that govern valid BT structural transformations. Under this dual-constraint framework, we employ a structured reflective experience format for precise BT plan modification:

\begin{equation}
e: \langle \tau,\ \omega \rangle \quad \text{where} \quad \omega = \underbrace{\text{"[OPERATION]"}}_{\text{core command}} + \underbrace{\text{"[RATIONALE]"}}_{\text{functional justification}}
\label{eq:instruction_schema}
\end{equation}

Specifically, the target stratum $\tau$ ($\alpha/\lambda/\Pi$) is determined by the preceding \textit{hierarchical plan analysis}, ensuring alignment with the identified plan flaw sources. The $\omega$ component integrates two essential elements: an imperative operation specifying concrete modifications (e.g., "Replace Sequence with Fallback") and a functional rationale explaining the modification intent (e.g., "this allows the drone to attempt different strategies"). This integrated approach not only ensures precision but also maintains semantic consistency with mission objectives, thereby establishing a correction paradigm that effectively balances accuracy with interpretability.

\begin{algorithm}[H]
\caption{Hierarchical BT Modification}
\label{alg:bt_modification}
\begin{algorithmic}[1]
\REQUIRE 
    $BT_{\text{init}}$: Initial LLM-generated BT \\
    $E_{\text{base}}$: Experience Base \\
    $\mathcal{A}$, $\mathcal{L}$: Action/Logic constraint spaces
\ENSURE 
    $E_{\text{new}}$: Reflective experiences for current iteration

\STATE $E_{\text{new}} \gets \emptyset$  \hfill \textit{// Initialize experience set}
\STATE $(\mathcal{F}_\alpha, \mathcal{F}_\lambda, \mathcal{F}_\Pi) \gets \text{HierarchicalAnalysis}(BT_{\text{init}})$  
\hfill \textit{// Single-step layer analysis}

\FOR{\textbf{each} layer $\tau \in \{\alpha, \lambda, \Pi\}$}
    \FORALL{flaw $f \in \mathcal{F}_\tau$}
        \STATE $\omega \gets \text{GenerateOp}(f, \mathcal{A}, \mathcal{L}) + \text{GenerateRation}(f)$  
        \hfill \textit{// Operation + rationale}
        \STATE $E_{\text{new}} \gets E_{\text{new}} \cup \{ \langle \tau, \omega \rangle \}$  
        \hfill \textit{// Structured reflective experience}
    \ENDFOR
\ENDFOR

\STATE $E_{\text{base}} \gets E_{\text{base}} \cup E_{\text{new}}$  
\hfill \textit{// Update experience repository}
\RETURN $E_{\text{new}}$  \hfill \textit{// Experiences for BT modification}
\end{algorithmic}
\end{algorithm}

\subsubsection{Experience Base}

Reflective experience $E$ can be directly utilized to modify the BT plan. Concurrently, we categorize experiences by task units and preserve each $E$ obtained through reflection during iterative cycles. This approach endows the Experience Base with transferability to arbitrary similar tasks, whereas individual-step $E$ is strictly bound to the plan snapshot of its generation iteration. We employ a kNN retriever with the \textit{all-mpnet-base-v2} model \cite{song2020mpnet} to retrieve $k$ most relevant task experiences. 

Owing to this architectural design, our methodology not only facilitates dynamic task plan adjustments but also enables acquisition of domain-specific knowledge through offline iterations. This knowledge can subsequently be transferred to real-world mission scenarios.

\section{Implementation}
\subsection{Overall Implementation}
\label{sec:implement}
\sysname employs an end-to-end system integration approach deployed on the real-world drone platform.

\textbf{Software Implementation:} 
The core system is primarily implemented in C++. Communication with the low-level flight controller is handled via MAVROS \cite{mavros}, while the behavior tree framework leverages the BehaviorTree.CPP library \cite{btcpp} for execution. Python components manage requests to remote server-based LLMs and operating the onboard vision module using the YOLOv5 object detection model \cite{YOLOv5}. 

\textbf{Hardware Platforms:} The experimental platform is ZHUOYI FS-J310 multirotor drone, equipped with an NVIDIA Jetson Orin NX onboard computer (16GB RAM) \cite{jetson}, integrating a 6-core Arm® Cortex®-A78AE v8.2 64-bit CPU and a 1024-core NVIDIA Ampere architecture GPU with 32 Tensor Cores. The hardware stack included a CUAV V5 Nano Autopilot flight controller \cite{cuav}, a Livox Mid-360 LiDAR \cite{lidarweb}, and a FASTLIO localization module \cite{FASTLIO}.
We use Intel D435i~\cite{d435i} as vision camera to obtain RGB image and depth estimation. Note that  
our system remains agnostic to the specific methods used for image acquisition or depth estimation—compatible alternatives include Edge YOLO~\cite{liang2022edge} for object detection, or stereo depth solutions like MobiDepth~\cite{zhang2022mobidepth} and AnyNet~\cite{wang2018anytime}.

\begin{figure*}
    \centering
    \includegraphics[width=1\linewidth]{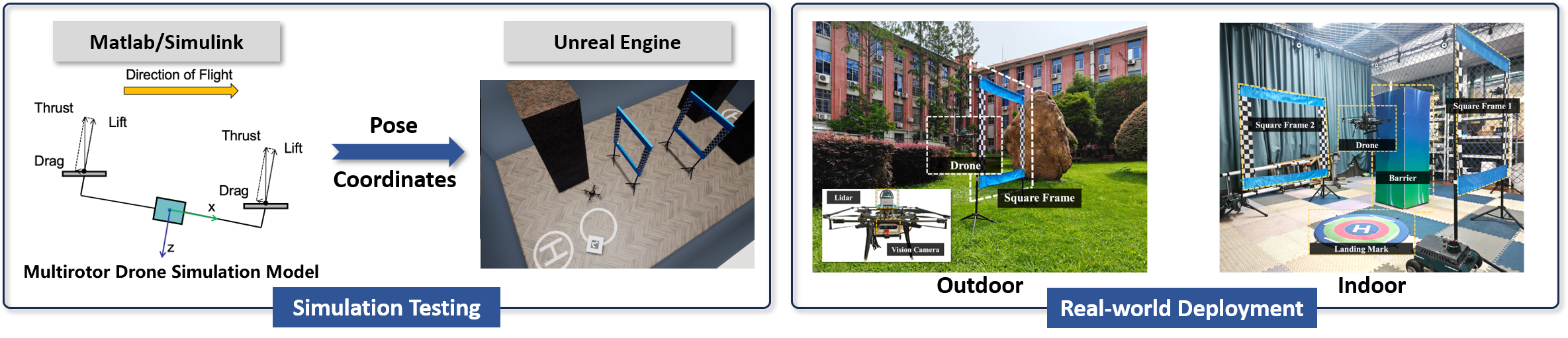}
    \caption{Experimental scenarios of \sysname.  Left: Simulation testing environment employing PX4 Software-in-the-Loop/Hardware-in-the-Loop drone models with real-time pose visualization via Unreal Engine; Right: Real-world deployment scenarios replicating identical configurations in both indoor and outdoor environments.}
    \label{fig:scenario}
\end{figure*}
\subsection{BT Plan Generation}
\label{sec:generator}
We use XML-formatted BTs as our plan representation. Given the strictly defined syntax of BTs, constraining the LLM's output format is essential. While some approaches employ fine-tuning \cite{izzo2024btgenbot} or complex algorithms \cite{10.24963/ijcai.2024/755} to generate BT plans as task planning results, we propose a simple one-shot method. Our approach remains agnostic to the underlying task planning process, focusing solely on output format constraints. This ensures compatibility with any reasoning framework.

We first provide the XML syntax specification and functional documentation for each node type, enabling the LLM to correctly interpret node capabilities. Subsequently, we supply a complete, simple Behavior Tree one-shot sample in the prompt. Empirically, we recommend incorporating syntax examples for blackboard parameter referencing, as this significantly improves reliability. 

\section{Experiment}
\subsection{Experimental Setup}

\textbf{Experimental Scenarios:} 

Figure~\ref{fig:scenario} illustrates our comprehensive experimental setup encompassing both simulation and real-world deployment. Simulation tests employed PX4 Software-in-the-Loop (SIL) and Hardware-in-the-Loop (HIL) environments for unmanned system control and flight dynamics modeling. Real-time pose (position and orientation) and coordinate data are visualized in 3D using Unreal Engine.

During real-world deployment, experiments are conducted across indoor and outdoor environments while monitoring onboard computational load. Given the high cost of mission failures in practical applications, we directly leveraged the validated experience base refined through simulation iterations to verify task success rates. The drone model FS-J310, as referenced in Section \ref{sec:implement}, is used for all field tests. 


\textbf{Benchmark:} Our experiments include four task types: three typical tasks for drone operations (\textit{Path Planning}, \textit{Object Searching}, \textit{Obstacle Navigation}) and \textit{Composite Task} that builds upon these core capabilities. Detailed task instructions and scene configurations for each type are provided in Table \ref{tab:task}.

\begin{table*}[t]
\centering
\renewcommand{\arraystretch}{1.5}
\caption{Task descriptions and scene configurations.}
\begin{tabularx}{\textwidth}{|c|c|X|X|}  
\hline
\rowcolor[HTML]{F5F5F5} \textbf{Type} & \textbf{ID} & \textbf{Task Instruction} & \textbf{Scene Configuration} \\  
\hline
\multirow{3}{*}{Path Planning} 
    & 1 & Fly forward first, then fly left to the target point.
    & Center point at (2, 2) \\ \cline{2-4}
    & 2 & Fly a 2x2 square path around the center point.
    & Center point at (1, 1) \\ \cline{2-4}
    & 3 & Avoid No-Fly zones and proceed to the target point
    & target (4,1), No-fly zones range in (1,0),(1,3),(2,0),(2,3) \\ \hline
    
\multirow{3}{*}{Object Searching} 
    & 4 & Find a balloon in the room 
    & The balloon is located at (-2, 3, 1) \\ \cline{2-4}
    & 5 & Locate the square frame 
    & Position of the square frame is randomly initialized\\ \cline{2-4}
    & 6 & Search for a Landing mark on the floor     
    & Landing mark position randomly initialized \\ \hline
    
\multirow{3}{*}{Obstacle Navigation} 
    & 7 & Fly over the cylinder at 1m height
    & Cylinder is located at (1,1) \\ \cline{2-4}
    & 8 & Navigate around the rectangular obstacle 
    & Obstacle is located at (1,0)\\ \cline{2-4}
    & 9 & Cross through the square frame 
    & Square frame is located at (1, 0)\\ \hline
    
\multirow{2}{*}{Composite Task}
    & 10 & Find and cross through the square frame 
    & Position of the square frame is randomly initialized\\ \cline{2-4}
    & 11 & Avoid No-Fly Zones and land on landing mark
    & No-fly zones range in (1,0),(1,3),(2,0),(2,3)\\ \hline
\end{tabularx}

\vspace{-0.3em}
\label{tab:task}
\end{table*}

\textbf{Baselines:}
We select representative frameworks from distinct planning paradigms: 
static task planning (\textit{ChatFly} \cite{chen2023typefly}), 
and self-updating task planning (\textit{LLM-Planner} \cite{song2023llmplanner} and \textit{REFLECT} \cite{liu2023reflect}). All baselines are obtained from their official repositories and adapted to our experimental setup. The standard LLM-Planner includes replanning; we also evaluate a modified variant, \textit{LLM-Planner-HLP}, which disables replanning to better assess its static planning capability.
Notably, while \textit{LLM-Planner} utilizes this kNN-based retrieval approach, all other baselines employ only one-shot examples.
Since \textit{REFLECT} lacks a dedicated planner design, we use the same planner described in Section \ref{sec:generator} as in \sysname.

\textbf{Metrics:} 
For planning performance, we measure Success Rate (\textit{\textbf{SR}}): the ratio of successfully completed user instructions. For iterative methods, we allow up to 5 refinement steps.  SR is calculated as:
\begin{equation}
\textit{\textbf{SR}} = \frac{ \sum_{i=1}^{5} N_{\text{success}}^{(i)} }{N_{\text{total}}}
\end{equation}
where $N_{\text{success}}^{(i)}$ is the number of tasks succeeding at the $i$-th iteration, and $N_{\text{total}}$ is the total number of tasks.

For task evaluation methods, we assess:
Detection (\textit{\textbf{Det}}): ratio of detected task failures to total occurrences, 
Localization (\textit{\textbf{Loc}}): precision in identifying the initial position of execution errors in the plan, and 
Explanation (\textit{\textbf{Exp}}): percentage of failure explanations deemed both reasonable and correct by human evaluators.

subsection{Overall Performance}
\subsubsection{Comparison with Baselines}

\begin{table*}[htbp]
    \centering
    \normalsize
    \caption{Quantitative comparison of \sysname and baseline methods. 
         All approaches evaluated without human-in-the-loop.
         \sysname achieves SOTA performance across all scenarios.}
    \begin{tabular}{l c c c c}
        \toprule
        \textbf{Method/Task} & \textbf{Path Planning} & \textbf{Object Searching} & \textbf{Obstacle Navigation} & \textbf{Composite Task} \\
        \midrule
        \textit{ChatFly} (2025 TMC) & 11.03\% & 18.46\% & 21.74\% & 3.57\% \\
        \textit{LLM-Planner-HLP} & 48.28\% & 30.43\% & 26.09\% & 7.69\% \\
        \textit{LLM-Planner} (2023 ICCV) & 48.28\% & 43.48\% & 26.09\% & 15.79\% \\
        \textit{REFLECT} (2023 CoRL) & 31.03\% & 52.17\% & 39.13\% & 26.32\% \\
        \midrule
        \rowcolor[HTML]{ECF4FF}  \sysname(Ours) & \textbf{84.91\%} & \textbf{88.46\%} & \textbf{91.67\%} & \textbf{80.35\%} \\             
        \bottomrule
    \end{tabular}    
    \label{tab:comparison}
\end{table*}

Table \ref{tab:comparison} presents the \textbf{\textit{SR}} evaluation of \sysname compared against SoTA baselines across four benchmark scenarios in the simulation environment. 
\sysname achieves the best performance across all benchmark scenarios with notable margins of improvement:
Path Planning: 84.91\% (at least improve 75.9\%),
Object Searching: 88.46\% (at least improve 69.6\%),
Obstacle Navigation: 91.67\% (at least improve 134.2\%),
Composite Task: 80.35\% (at least improve 205.3\%).
We next analyze the underlying mechanisms for \sysname's significantly superior performance by combining insights from the unique characteristics of different task scenarios.


When handling tasks without explicit failure signals, such as the \textit{Path Planning} task requiring assessment of intermediate waypoints during drone flight, baseline methods like \textit{LLM-Planner} (relying on explicit failure signals) or \textit{REFLECT} (employing only final-state verification) prove ineffective. In contrast, \sysname with \textit{Continuous State Evaluation} mechanism enables assessment of dynamic processes, providing robust outcome determination and failure explanation to drive iterative optimization.

For tasks demanding structurally complex logical control flows, particularly the \textit{Composite Task} scenario requiring BT syntactic constraints during exceptions like undetected target objects, existing coarse reflection methods accumulate unstructured experience, resulting in inefficient iterative refinement. \sysname's \textit{Hierarchical BT Modification} approach overcomes this limitation through precise hierarchical adjustments of BT plans, thus achieving superior refinement outcomes.

\subsubsection{Deployment in Real World}

\begin{figure}[htbp] 
  \centering
  \subfloat[Path planning flight trajectory (ID 2)]{%
    \label{fig:pathplan}%
    \includegraphics[width=0.47\linewidth]{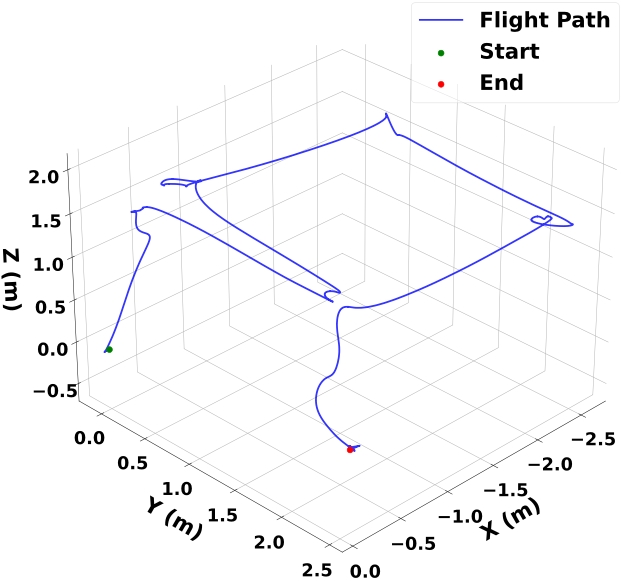}
  }%
  \hfill 
  \subfloat[Composite task flight trajectory (ID 10)]{%
    \label{fig:composite}%
    \includegraphics[width=0.47\linewidth]{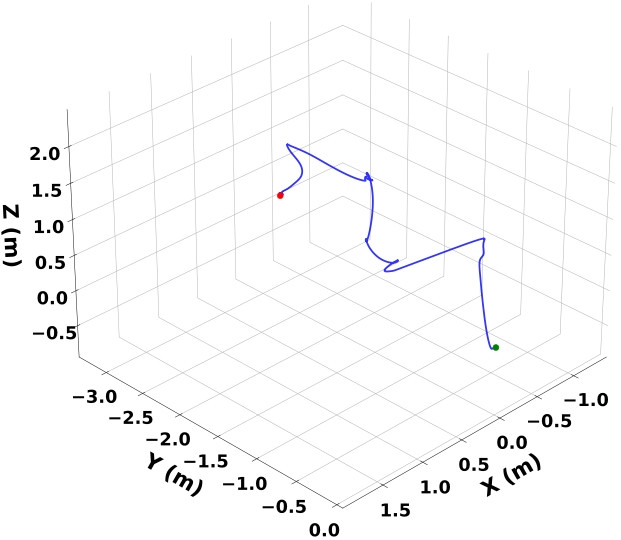}
  }%
  \caption{Flight path 3D visualization. (a) The drone executes a 2m×2m square trajectory around a central point in the XY plane. (b) Autonomous navigation demonstration where the drone sequentially detects and passes through two square frames positioned along its flight path.}
  \label{fig:trajectory}
\end{figure}

This section presents experimental validation of the \sysname framework through real-world deployment in physical environments, with each task performed ten times indoors and ten times outdoors. The primary objectives are to verify the system's practical effectiveness in autonomous aerial tasks and demonstrate its feasibility for deployment on main-stream drone platforms with limited computational resources. Physical testing was exclusively conducted with our refined experience base due to unacceptable physical risks associated with trial-and-error learning in real-world settings, where task failures could incur high-cost damages (e.g., sensor destruction or personnel injury). To provide visual insights into specific task execution patterns, we present 3D flight path visualizations of Task 2 and Task 10 in Figure~\ref{fig:trajectory}.

Analysis of experimental results in Table~\ref{tab:validation_results} first reveals consistently high success rates across environments, 95\% indoor and 97.5\% outdoor, demonstrating \sysname's operational effectiveness regardless of environment. 
Detailed failure analysis further establishes that all observed errors originate exclusively at the execution level. 
Specifically in Task 4, the single failure occurred during balloon identification due to perception system limitations, while Task 9 and Task 10 failures stemmed from environmental misidentification of navigation frames. 


\begin{figure*}[t]
    \centering
    \subfloat[CPU Workload\label{fig:cpu}]{
        \includegraphics[width=0.31\linewidth]{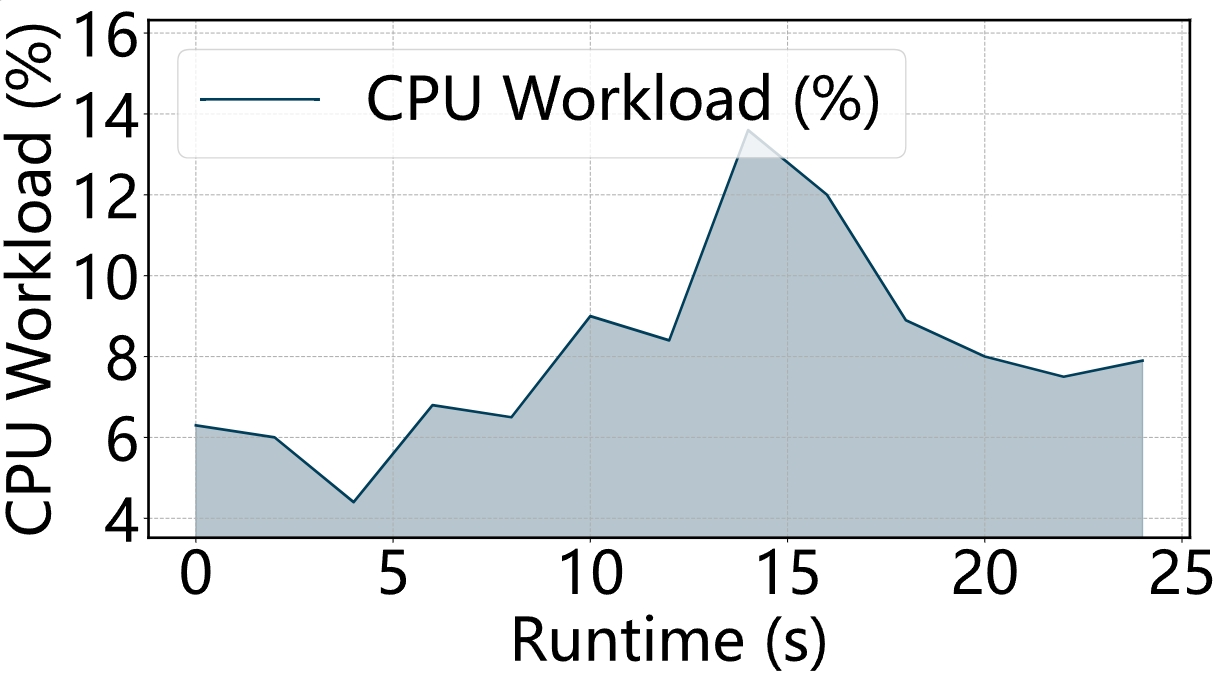}  
    }
    \hfill
    \subfloat[Total Power\label{fig:power}]{
        \includegraphics[width=0.31\linewidth]{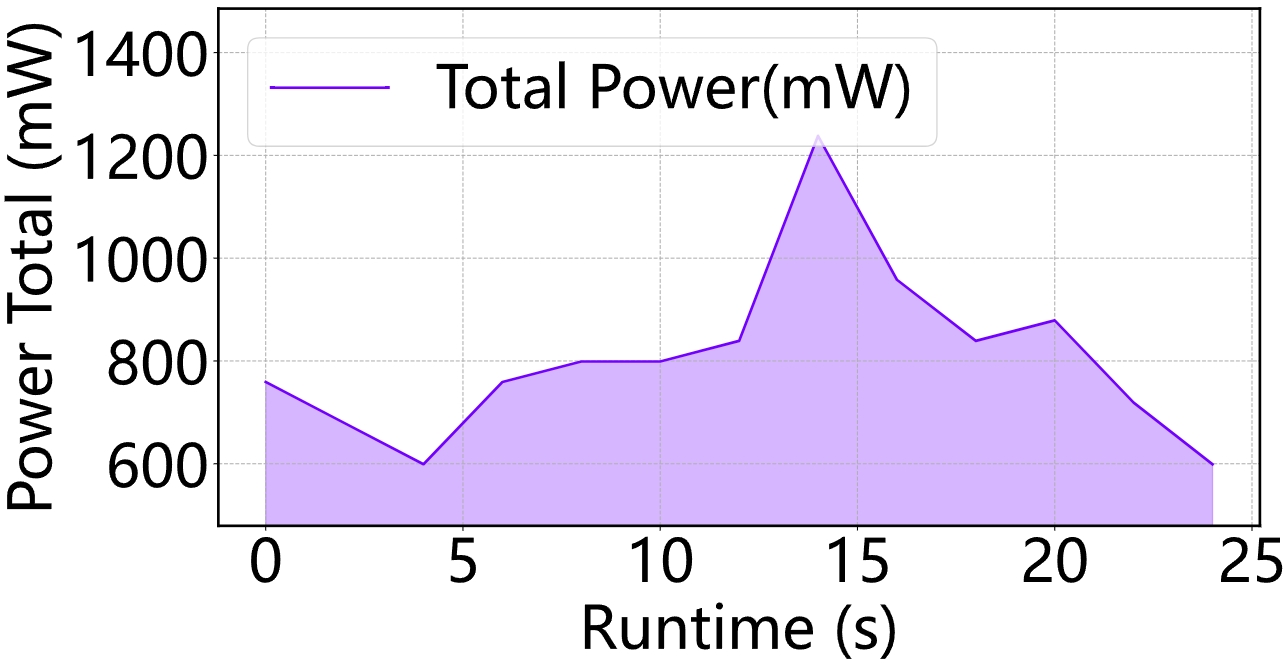}
    }
    \hfill
    \subfloat[RAM Usage\label{fig:ram}]{
        \includegraphics[width=0.31\linewidth]{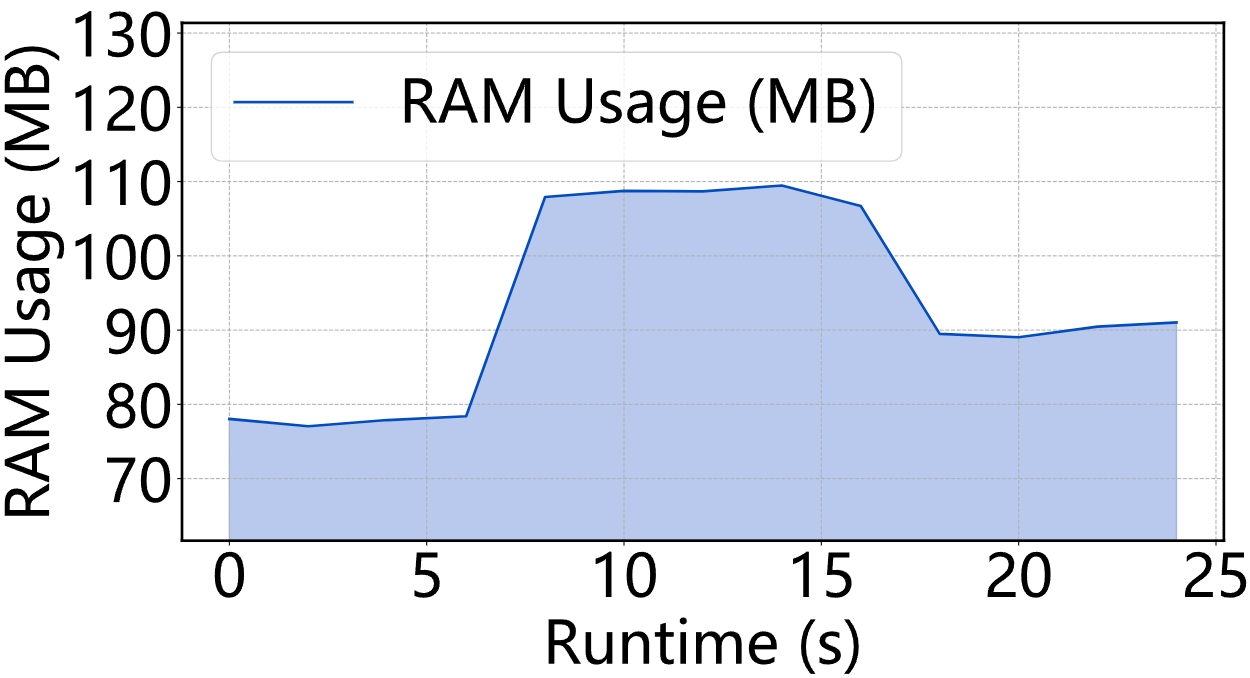}
    }
    \caption{Performance of \sysname\ on physical drone platform.}
    \label{fig:deployment}
\end{figure*}

\begin{table}[ht]
    \centering
    \caption{Real-world deployment experiments.}
    \renewcommand{\arraystretch}{1.5} 
    \begin{threeparttable}
    \begin{tabular}{|c|c|c|}
        \hline
        \rowcolor[HTML]{F5F5F5} \textbf{ID\tnote{$*$}} & \textbf{Indoor Success/Total (Rate)} & \textbf{Outdoor Success/Total (Rate)} \\
        \hline
\multicolumn{3}{c}{} \\[-2ex]  
\hline
      2 & 10/10 (100\%) & 10/10 (100\%) \\
        \hline 
        4 & 10/10 (100\%) & 9/10 (90\%) \\
        \hline 
        9 & 9/10 (90\%) & 10/10 (100\%) \\
        \hline 
        10 & 9/10 (90\%) & 10/10 (100\%) \\
        \hline
    \end{tabular}
    \begin{tablenotes}
    \footnotesize
    \item [$*$]Detailed descriptions of ID numbers are provided in Table~\ref{tab:task}.
    \end{tablenotes}
    \end{threeparttable}
    \label{tab:validation_results}
\end{table}

To further validate the practical feasibility of \sysname, we conducted comprehensive measurements of key resource consumption metrics during real-time operation. As shown in Figure~\ref{fig:deployment}, the system maintains peak CPU overhead below 14\%, RAM usage under 110 MB, and total power consumption within 1300 mW during task execution. These results are obtained through continuous monitoring of ZHUOYI FS-J310 multirotor drone (NVIDIA Jetson Orin NX \cite{jetson}). The low runtime overhead highlights that our system imposes minimal load on the device, ensuring ease of deployment in real-world scenarios.

The system's overhead shows clear variations during different stages of task execution. The 12-16 second period shows increased CPU and memory usage, corresponding to the activation of visual perception modules for scene understanding. The slightly elevated load in the final phase is attributed to the \textit{Continuous State Evaluation} module, which involves data stream filtering, state logging operations, and communication with cloud-based LLM services. These transient increases remain well within the hardware's operational limits, demonstrating the system's robustness under dynamic workload variations.

\subsection{Performance Breakdown}
We next evaluate the performance of the key components of the \sysname in detail.
\subsubsection{Plan-level Failure Detection and Explanation}

Plan-level failures occur when a drone completes its action sequence but fails to achieve the task goal, making them more complex to diagnose than execution failures.
Therefore, we compare our proposed \textit{Continuous State Evaluation} method with the \textit{Final State-based} method and an ablated version of our model without CMSR (\textit{Ours w/o CMSR}).  All comparisons are conducted on the collected 176 plan-level failures from both simulation and real-world drone task executions.

Experimental results in Table~\ref{tab:failure} indicate that \textit{Final State-based} method suffers severe limitations in detecting plan-level failures due to its inability to model task processes, resulting in poor performance (\textit{Det}=23.08\%, \textit{Exp}=14.72\%, \textit{Loc}=12.18\%). 
\textit{Ours w/o CMSR} leverages task process data streams for evaluation, achieving moderately higher results (\textit{Det}=31.41\%, \textit{Exp}=20.83\%, \textit{Loc}=18.59\%) than \textit{Final State-based}. However, it remains significantly inferior to \sysname due to the LLM's limitations in processing time-serial data.

\sysname with \textit{Continuous State Evaluation} framework achieves superior failure analysis quality (\textit{Det}=90.38\%, \textit{Exp}=85.26\%, \textit{Loc}=80.07\%). This result underscores the necessity of process state assessment for drone task scenarios. Final-state evaluations fail to capture systemic planning issues, whereas our approach effectively addresses these challenges by leveraging the semantic continuity of the task process. Thus, our system provides a robust solution for plan-level failure detection and explanation in drone mission scenarios.


\begin{table}[h]
\centering
\normalsize
\caption{Plan-Level failure evaluation comparison.}
\begin{tabular}{lcccc}
\toprule
\textbf{Methods} & \textbf{Det} & \textbf{Exp} & \textbf{Loc} \\
\midrule
Ours w/o CMSR & 31.41 & 20.83 & 18.59  \\
Final State-based & 23.08 & 14.72 & 12.18  \\
\midrule
\rowcolor[HTML]{ECF4FF} \textbf{\sysname(Ours)} & \textbf{90.38} & \textbf{85.26} & \textbf{80.07} \\
\bottomrule
\end{tabular}
\label{tab:failure}
\end{table}

\subsubsection{Iterative Refinement Process Comparison}

\begin{figure}[htbp]
    \centering
    \subfloat[\small Path Planning]{    
        \begin{minipage}{0.47\columnwidth}
            \centering
            \includegraphics[width=\linewidth]{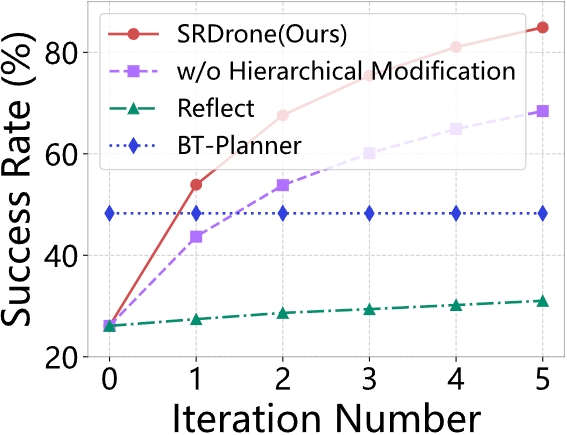}
            \label{fig:sub1}
        \end{minipage}
    }%
    \hfill
    \subfloat[Object Searching]{
        \begin{minipage}{0.47\columnwidth}
            \centering
            \includegraphics[width=\linewidth]{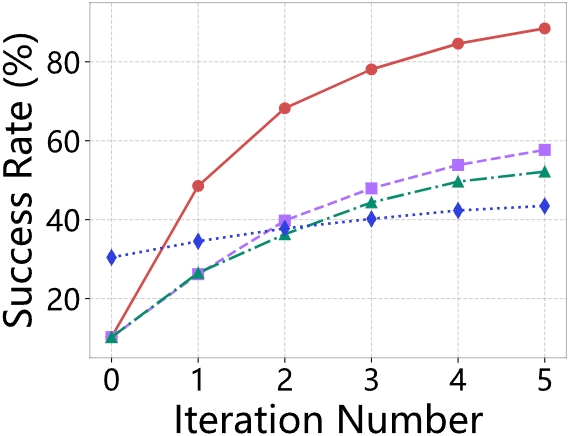}
            \label{fig:sub2}
        \end{minipage}
    }

    \vspace{0.5em} 

    \subfloat[\small Obstacle Navigation]{
        \begin{minipage}{0.47\columnwidth}
            \centering
            \includegraphics[width=\linewidth]{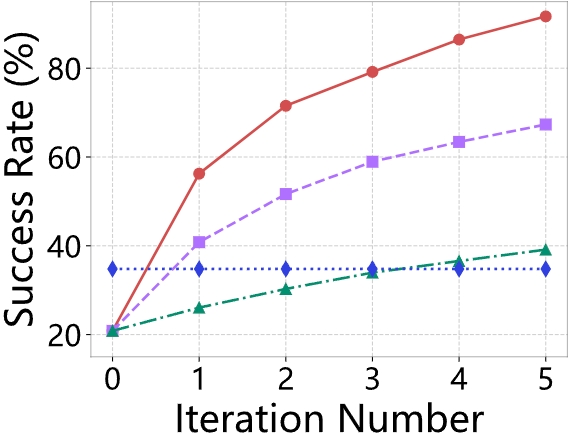}
            \label{fig:sub3}
        \end{minipage}
    }%
    \hfill
    \subfloat[\small Composite Task]{
        \begin{minipage}{0.47\columnwidth}
            \centering
            \includegraphics[width=\linewidth]{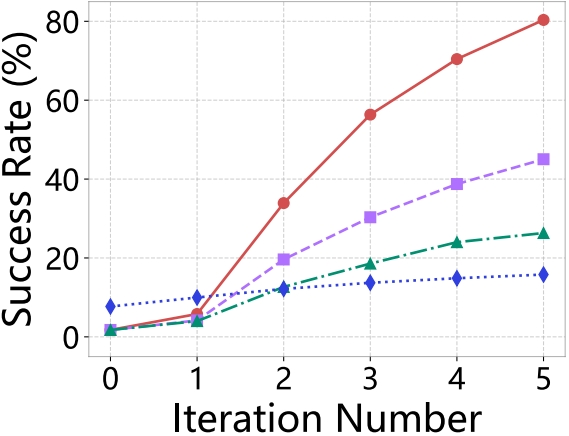}
            \label{fig:sub4}
        \end{minipage}
    }
    \caption{\small Iterative refinement process comparison.}
    \label{fig:refine}
\end{figure}

Figure \ref{fig:refine} illustrates the iterative refinement process, featuring our \sysname system alongside the \textit{REFLECT} and \textit{LLM-Planner} baselines, both equipped with dynamic planning adjustment capabilities. We also include an ablation study group ("\textit{w/o Hierarchical Modification}") which removes the entire Hierarchical Modification module, directly utilizing failure explanations from the Evaluation module for BT Modification.



The experimental results establish \sysname's significant superiority over \textit{REFLECT} under identical initial plans. Notably, while \textit{LLM-Planner} achieves higher initial success rates due to its advanced planner, \sysname surpasses this baseline within two refinement iterations. As optimization cycles progress, \sysname demonstrates increasingly dominant performance over the ablation variant \textit{w/o Hierarchy Modification}, conclusively validating the efficacy of our \textit{Hierarchical BT Modification} module.

While both \textit{REFLECT} and the ablation variant \textit{w/o Hierarchical Modification} achieve comparable result due to near-ubiquitous fault detection, the ablation's slight performance advantage originates directly from \sysname's higher-quality task process evaluation. This superior assessment capability enables more precise failure diagnosis even without hierarchical modification, demonstrating clear advancement over \textit{REFLECT}'s failure explanation mechanism.

For tasks requiring architectural management of multi-layered BT structures, our \textit{Hierarchical BT Modification} enables fine-grained plan adjustments. The \textit{Composite Task} scenario exemplifies this challenge, demanding concurrent syntactic satisfaction and exception handling within complex BT architectures. In such environments, \sysname maintains significant performance advantages over the \textit{w/o Hierarchical Modification} ablation. The widening performance gap with increasing plan complexity conclusively demonstrates the critical role of hierarchical refinement mechanisms.

\begin{table*}[htbp]
    \centering
    \normalsize
    \caption{Quantitative comparison of \sysname and Human-Guide variants.}
    \begin{tabular}{l c c c c}
        \toprule
        \textbf{Method/Task} & \textbf{Path Planning} & \textbf{Object Searching} & \textbf{Obstacle Navigation} & \textbf{Composite Task} \\
        \midrule        
        General-Guide & 86.82\% & 80.21\% & 84.19\% & 68.44\% \\        
        Expert-Guide & 90.33\% & 92.14\% & 93.87\% & 82.50\% \\  
        \midrule
        \rowcolor[HTML]{ECF4FF} \sysname & 84.91\% & 88.46\% & 91.67\% & 80.35\% \\        
        \bottomrule
    \end{tabular}    
    \label{tab:humanloop}
\end{table*}

\subsubsection{Human-in-the-Loop Performance Comparison}
\label{sec:human_loop}

In addition to automated approaches, we introduce two human-in-the-loop configurations that substitute the system’s \textit{Continuous State Evaluation} and \textit{Hierarchical BT Modification} modules. In these configurations, humans directly oversee task execution and provide iterative plan refinement through natural language feedback. In the \textit{Expert-Guide} condition, participants possess expertise in AI and drone systems, whereas \textit{General-Guide} participants have no formal training in computer science or robotics. The quantitative comparison results are summarized in Table~\ref{tab:humanloop}.

\sysname achieves consistent superiority over \textit{General-Guide} across all scenarios except for \textit{Path Planning}.
This advantage  primarily stems from \sysname's structural reflective experience,  which enabling precise localization of BT plan flaws. 
In contrast, \textit{General-Guide} is hindered by the arbitrary mapping between high-level human preferences and low-level adjustments, often resulting in infeasible corrections that exceed the drones' physical constraints. In the \textit{Path Planning} scenario, \sysname exhibits slightly lower effectiveness compared to \textit{General-Guide}, a limitation attributable to the inherent challenges LLMs face in mathematical reasoning, even when compared to general users.

While \sysname shows a slight performance gap compared to \textit{Expert-Guide}, this discrepancy primarily stems from the latter’s ability to holistically integrate visual, spatial, and sensor data rather than relying on text-based abstractions, enabling more accurate error localization. Additionally, human experts surpass LLMs in geometric reasoning and trajectory optimization, allowing them to directly detect spatial constraint violations without requiring the iterative refinements necessary for autonomous systems.
Notably, the gap between \sysname and \textit{Expert-Guide} narrows most significantly in the \textit{Composite Task} scenario. This observation suggests that \sysname’s \textit{Hierarchical BT Modification} mechanism provides increasing benefits as task complexity grows, leveraging structured reflective experience to systematically resolve flaws across multiple levels of abstraction, thereby outperforming verbal feedback-based approaches.

\section{Related Work}


\textbf{Static Planning Methods}. Current approaches leverage Large Language Models (LLMs) to generate structured task plans using formalisms such as Python programs \cite{liang2023code} and Behavior Trees \cite{10.24963/ijcai.2024/755, ao2024llm}. These methods benefit from the expressiveness and industrial compatibility of such representations. Techniques including fine-tuning \cite{izzo2024btgenbot} and Reinforcement Learning \cite{ahn2022can} are further employed to enhance LLMs' embodied physical reasoning capabilities, ultimately aiming to increase the task success rate of generated plans.
However, these static planning methods exhibit limitations in dynamic adaptation. They produce fixed executable specifications which lack online adaptation mechanisms. This characteristic results in performance degradation when encountering OOD scenarios such as unfamiliar settings or unexpected environmental conditions. Consequently, the inherent rigidity of static planners fundamentally constrains autonomous agents' capacity to respond to real-world uncertainties.

To solve these limitations, \sysname introduces the Hierarchy Plan Modification mechanism, which enables precise adjustments to BT plans during execution and overcomes the rigidity of static planners. Furthermore, by leveraging the LLM's self-reflection capabilities, our method does not require pre-training on specific scenarios, thereby enhancing scalability and robustness in out-of-distribution environments.

\textbf{Human Robot Collaboration}.
Recent studies have explored diverse strategies for integrating human expertise into LLM-based task planning autonomous systems. For instance, Chen et al.'s work~\cite{chen2023typefly}, where human experts define replanning conditions for dynamic environments, and the system leverages an LLM to trigger robust task re-planning when these conditions are met. Parakh et al.'s work~\cite{parakh2024lifelong} introduces a framework where linguistic instructions from experts guide the hierarchical construction of complex skills from low-level primitives, enabling task adaptation. Meanwhile, Liu et al.'s work~\cite{liu2024enhancing} adopts a tele-operation system that bypasses autonomous algorithms entirely, allowing experts to directly control robots for improved manipulation precision. In contrast, Zhao et al.'s work~\cite{zhao2025beyondexpert} demonstrates a learning paradigm where policies are distilled from minimal expert demonstrations, achieving superhuman performance through the DOUBLE EXPLORATION algorithm. Zha et al.'s work~\cite{zha2024distilling} further advances this field by proposing DROC, an LLM-based system capable of processing arbitrary language feedback to extract generalizable knowledge from corrections. Despite these innovations, a critical limitation persists: over-reliance on human expertise. 
Most approaches require users with domain specialized knowledge, rendering them inaccessible to general users~\cite{bommasani2021opportunities,sanh2022multitask}. This dependency not only contradicts the core design objective of drones—to reduce human workload—but also hinders scalability in real-world applications.

In \sysname, we design the \textit{Continuous State Evaluation} framework, which automates drone mission monitoring by replacing human experts with a continuous state evaluation pipeline. Specifically, the system automatically identifies mission-critical events and infers high-level task objectives through heuristic reasoning that leverages the foundational reasoning capabilities of LLMs, without requiring scenario-specific training. This not only enhances scalability in OOD environments but also enables real-time, self-explanatory decision-making through natural language insights.
\section{Discussion}

\textbf{Adaptability of \sysname:}
\sysname enables truly autonomous task execution without human expertise through dual adaptive mechanisms. The framework leverages fundamental behavior tree architectural principles to implement Hierarchy Fine-grained BT Modification, ensuring universal compatibility with any behavior tree-formatted plans regardless of task requirements. This capability provides inherent task-agnostic adaptability. Furthermore, it utilizes continuous state evaluation addressing core motion characteristics of drone operations, delivering agent-agnostic performance for heterogeneous drone fleets. These synergistic mechanisms drive a self-reflective iteration process that eliminates human supervision dependency. \sysname also maintains broad LLM compatibility with models like GPT~\cite{achiam2023gpt}, Gemini~\cite{team2023gemini}, DeepSeek~\cite{guo2025deepseek} and Qwen~\cite{bai2023qwen} by constraining only planner output formats while preserving native reasoning processes, enabling flexible deployment across diverse autonomous systems.

\textbf{The worst performance case:}
\sysname demonstrates strong performance in managing \textit{Plan-level} tasks but lacks the ability to provide detailed error diagnostics at the \textit{Execution-level}. For example, in the \textit{Object Searching} task, the \textit{Detect} action is often paired with \textit{TurnLeft} or \textit{TurnRight} to adjust the field of view. If the turning angle exceeds optimal limits, the drone may trigger the \textit{Detect} action before stabilizing, causing blurred images and subsequent detection failures. To mitigate this, we introduce an \textit{Action-Centric State Filtering} mechanism that discards images captured during unsuccessful \textit{Detect} actions, effectively reducing communication overhead and computational load. However, this design inherently limits the system’s ability to provide fine-grained failure explanations in such edge cases.

\textbf{Future Works:} 
\sysname currently relies on cloud-based LLMs requiring server communication, 
utilizes SOTA LLMs' context capabilities without dedicated experience base optimization by feeding raw historical data directly into prompts, 
and operates without visual-language integration. Future enhancements will focus on: 
(1) Developing edge-compatible LLMs for on-device deployment to eliminate cloud dependency; 
this aims to remove reliance on stable communication links and enable operational capabilities in adversarial environments.
(2) Implementing structured experience compression and retrieval mechanisms to transcend current context-window limitations; 
this enhancement will strengthen complex task learning capabilities through optimized knowledge distillation, while improving generalization performance across similar yet distinct mission scenarios.
(3) Integrating Vision-Language Models to enable cross-modal scene understanding for complex spatial decision-making scenarios; 
this integration will enhance spatial reasoning accuracy through multi-modal perception fusion, particularly for navigation and object interaction tasks in dynamic environments.

\section{Conclusion}
This paper presents \sysname, a self-refining framework for autonomous drone task planning and execution. The approach combines continuous state evaluation during mission runtime with hierarchical, fine-grained behavior tree (BT) plan modification, enabling adaptive and robust performance. \sysname achieves state-of-the-art results, delivering the highest success rates across diverse operational scenarios and exhibiting substantial performance gains in complex composite environments compared to existing methods. Furthermore, the effectiveness of \sysname is validated through extensive real-world experiments on physical drone platforms, demonstrating its practicality and reliability under realistic conditions.

\bibliographystyle{IEEEtran}
\bibliography{ref}

\begin{IEEEbiography}[{\includegraphics[width=1in,height=1.25in,clip,keepaspectratio]{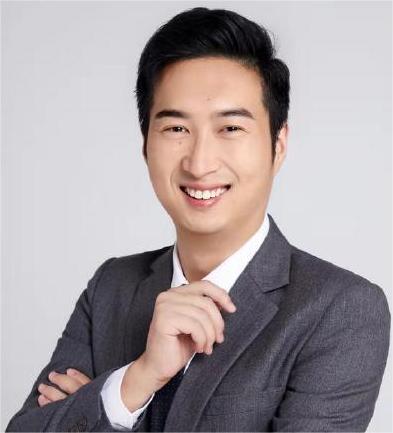}}]
{Deyu Zhang[S'14, M'16]}  received the B.Sc. degree in communication engineering from PLA Information Engineering University, Zhengzhou, China, in 2005, and the M.Sc. degree in communication engineering and Ph.D. degree in computer science from Central South University, Changsha, China, in 2012 and 2016, respectively. He is currently an Associate Professor with the School of Computer Science and Technology. His current research interests include mobile system optimization, edge computing, and stochastic optimization. He is a member of the ACM, IEEE, and CCF.
\end{IEEEbiography}

\begin{IEEEbiography}
[{\includegraphics[width=1in,height=1.25in,clip,keepaspectratio]{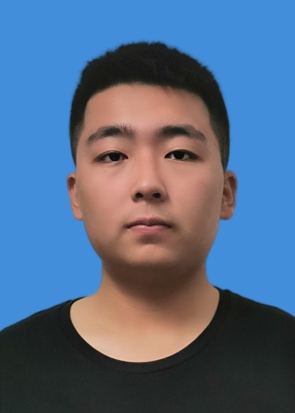}}]{Xicheng Zhang}
 received his B.Sc. degree in Computer Science from Central South University, China, in 2023. He is currently in the second year of his Ph.D studies in Computer Science at the same institution. His research interests include mobile computing and embodied intelligence.
\end{IEEEbiography}

\begin{IEEEbiography}
[{\includegraphics[width=1in,height=1.25in,clip,keepaspectratio]{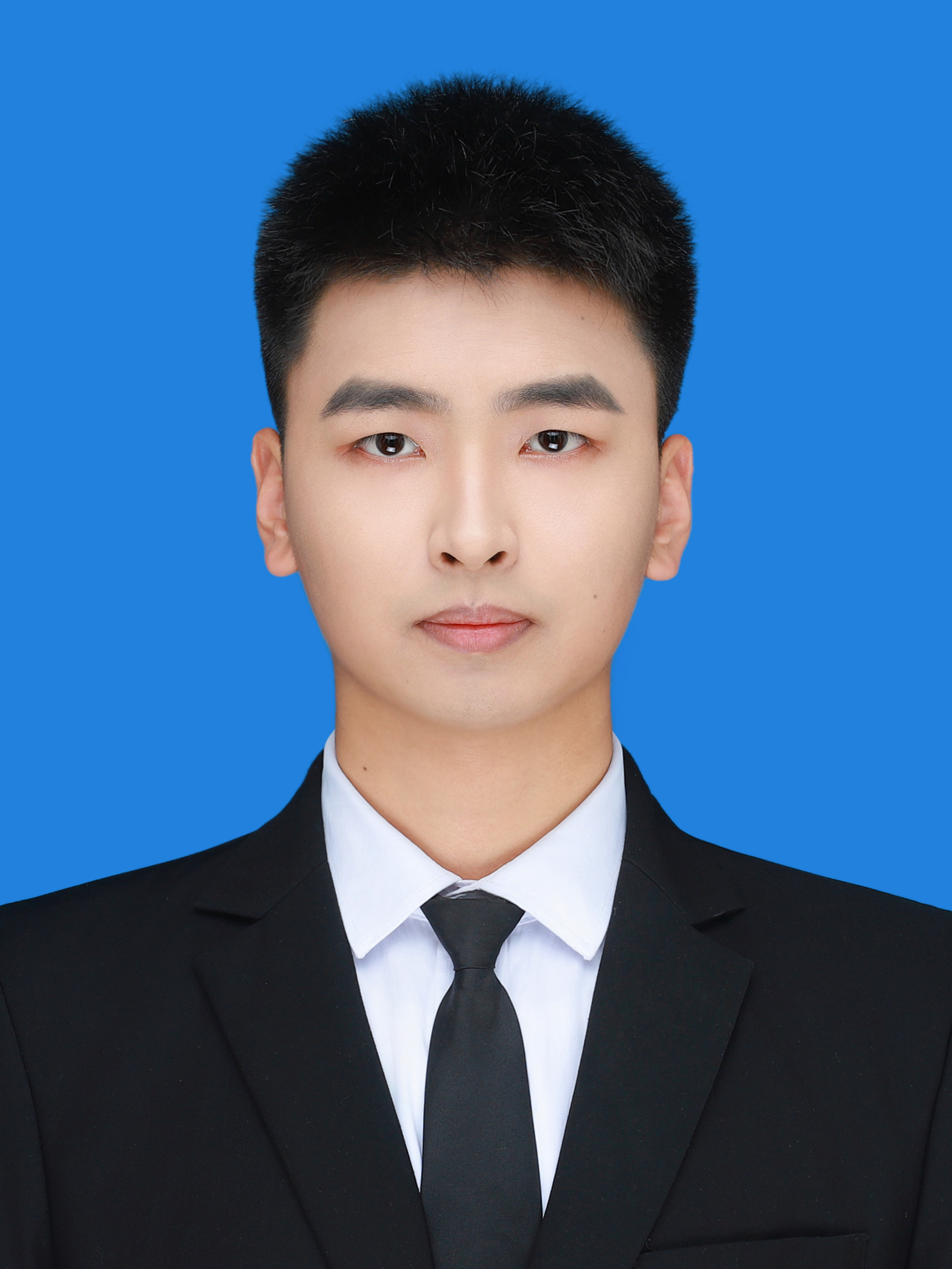}}]{Jiahao Li}
received his B.Sc. degree in Computer Science from Central South University, China, in 2024. He is currently in the second year of his M.Sc. studies in Computer Science at the same institution. His research interests include mobile computing and edge intelligence.
\end{IEEEbiography}

\begin{IEEEbiography}[{\includegraphics[width=1in,height=1.25in,clip,keepaspectratio]{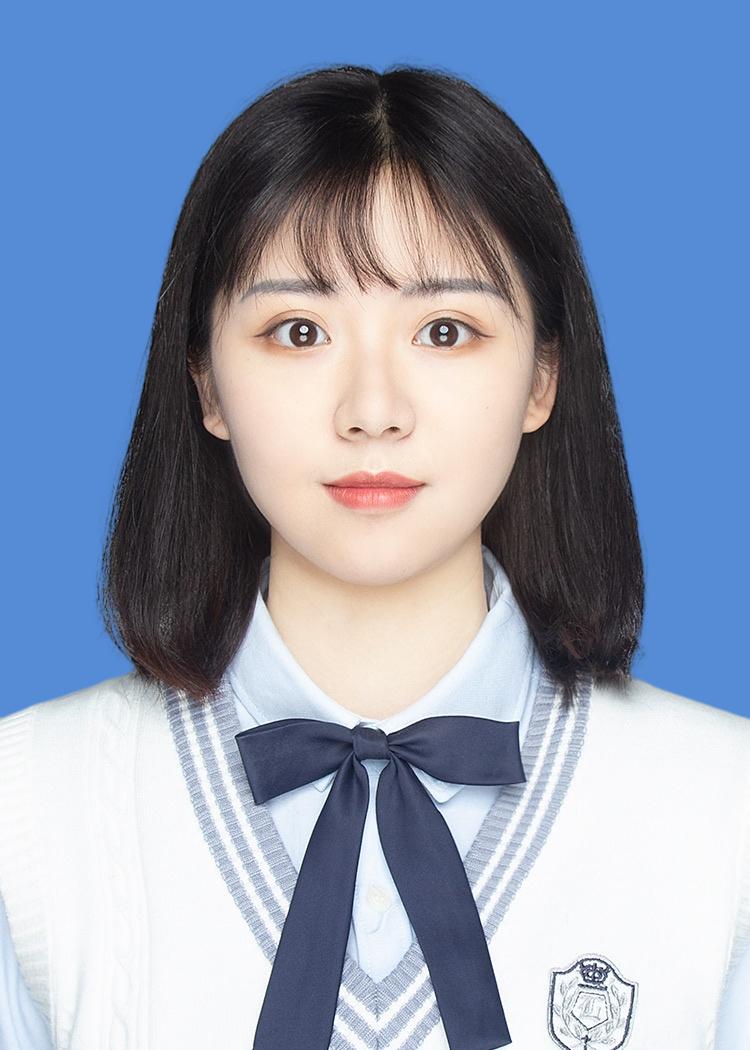}}]{Tingting Long}
 received her B.Sc. degree in Computer Science from Central South University, China, in 2023. She is currently in the second year of her M.Sc. studies in Computer Science at the same institution. Her research interests include mobile computing and edge intelligence.
 \end{IEEEbiography}

\begin{IEEEbiography}
[{\includegraphics[width=1in,height=1.25in,clip,keepaspectratio]{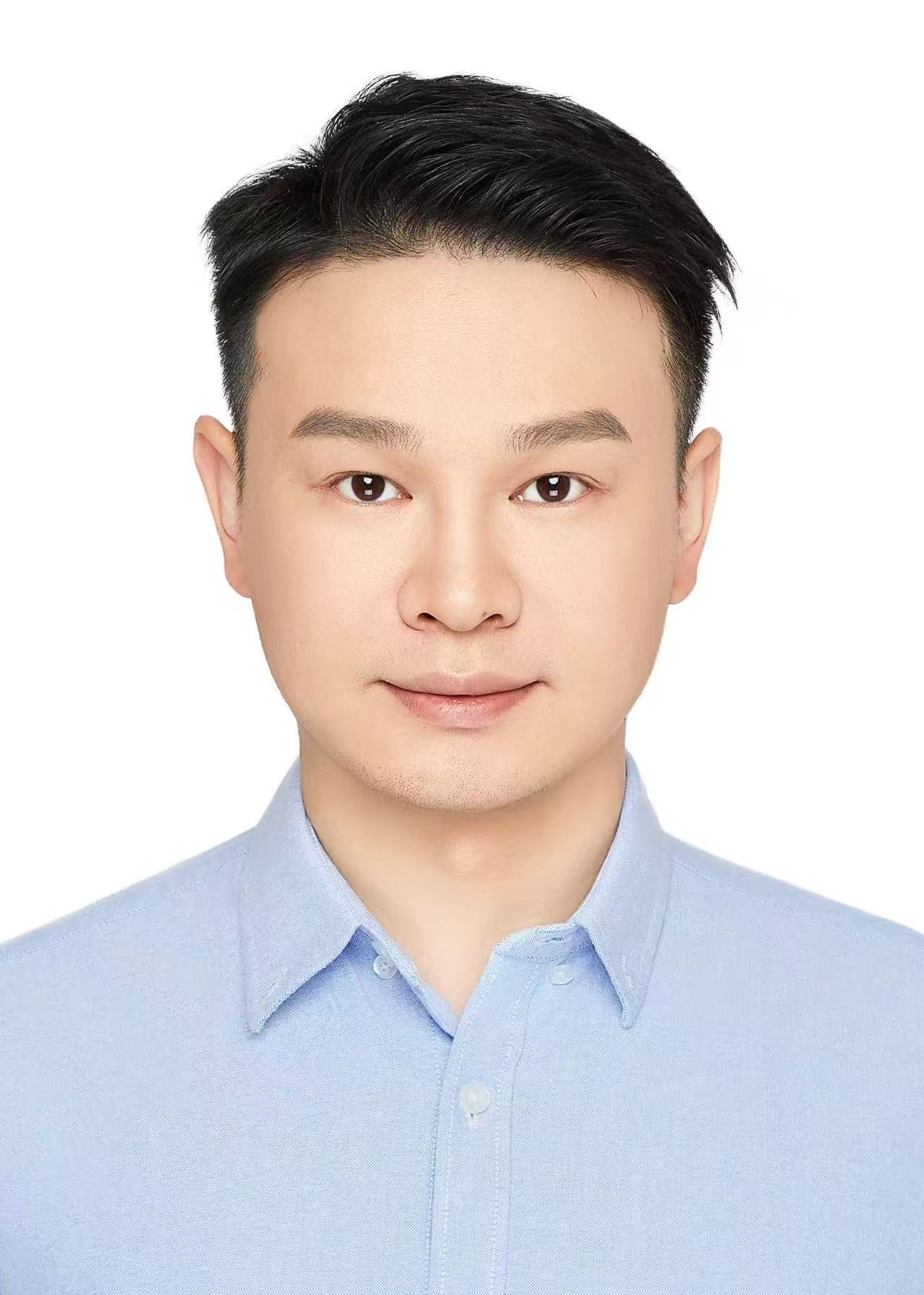}}]{Xunhua Dai}
received the B.S., M.S., and Ph.D. degrees in control science and
engineering at Beihang University, Beijing, China, in 2013, 2016,
and 2020, respectively.

Since 2020, he has been an Associate Professor with Central South
University in computer science and engineering, where he is currently
with the School of Computer Science and Engineering. His main research
interests include reliable intelligent control, safety assessment,
and design optimization of unmanned aerial robotics.
\end{IEEEbiography}

\begin{IEEEbiography}[{\includegraphics[width=1in,height=1.25in,clip,keepaspectratio]{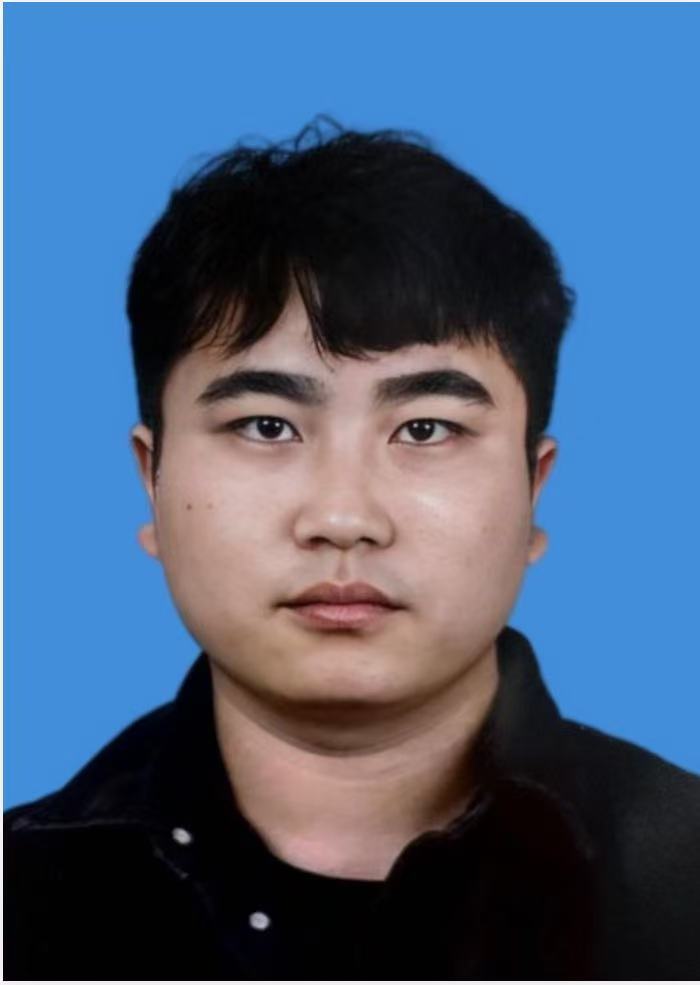}}]{Yongjian Fu}
received the B.Sc. in Computer Science from Central South University in 2021, China. He is currently working toward a Ph.D. degree in Computer Science at the School of Computer Science and Engineering from Central South University. His research interests include wireless sensing, mobile computing, and Internet-of-Things. \end{IEEEbiography}

\begin{IEEEbiography}[{\includegraphics[width=1in,height=1.25in,clip,keepaspectratio]{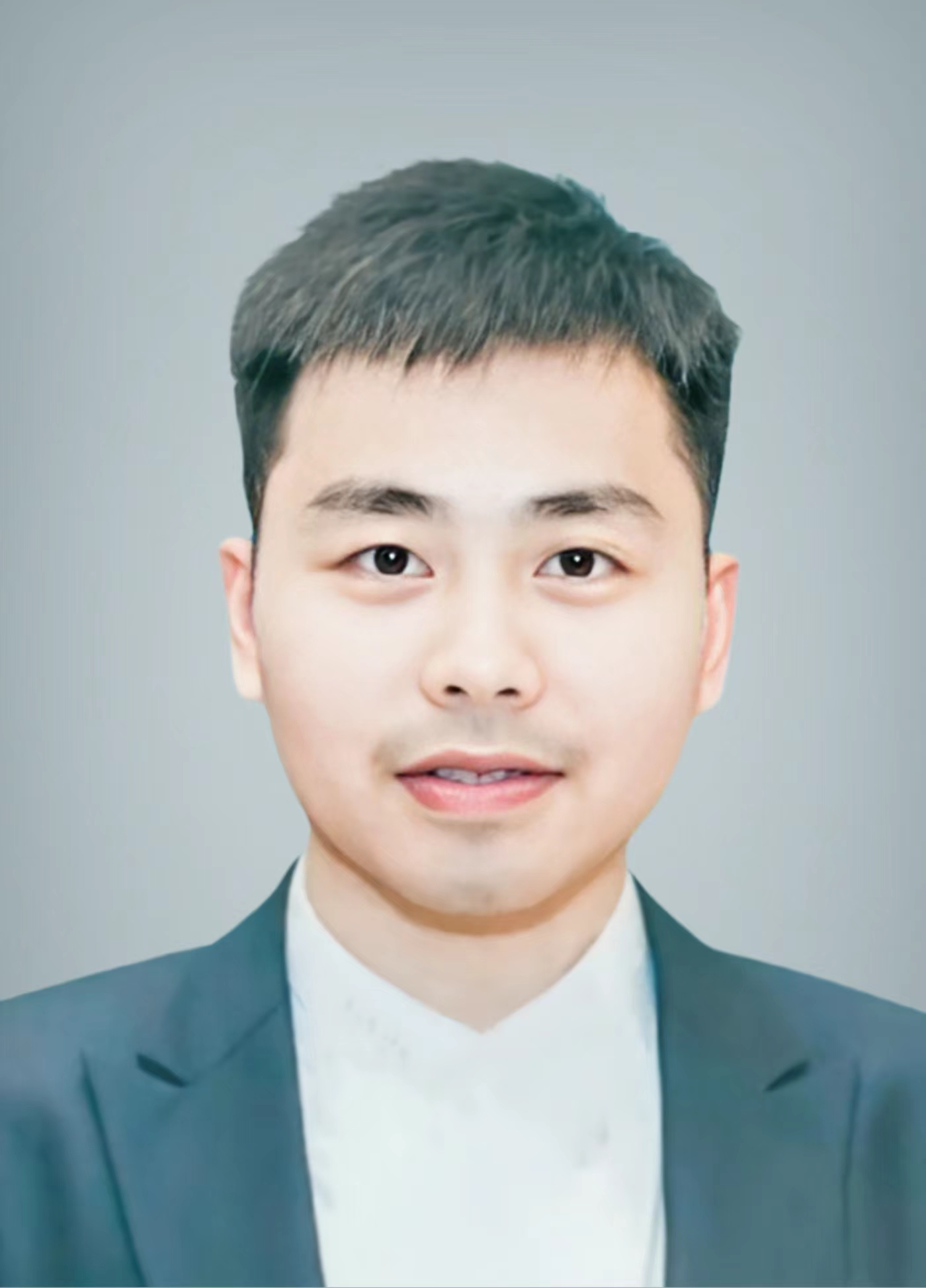}}]{Jinrui Zhang[S'19, M'23]}
received his B.Sc., M.Sc., Ph.D. degrees all in computer science, from Central South University, China, in 2016, 2018 and 2023, respectively. Currently, he is a postdoctoral researcher with the Department of Computer Science and Technology, Tsinghua University, China. He has worked as a visiting scholar at Seoul National University, Korea. His research interests include mobile computing, edge intelligence, and computer vision. He is a member of the ACM, IEEE, and CCF. \end{IEEEbiography}

\begin{IEEEbiography}[{\includegraphics[width=1in,height=1.25in,clip,keepaspectratio]{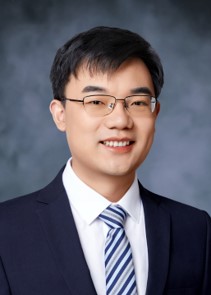}}]{Ju Ren[M'16,SM'21]}
 received his B.Sc., M.Sc., Ph.D. degrees all in computer science, from Central South University, China, in 2009, 2012 and 2016, respectively. Currently, he is an associate professor with the Department of Computer Science and Technology, Tsinghua University, Beijing, China. Prior to joining Tsinghua, he was a professor with the School of Computer Science and Engineering, Central South University, Changsha, China. his research interests include Internet-of-Things, edge computing, distributed \& embedded AI, and operating system. He is a senior member of the IEEE and a member of the ACM. He was recognized as a highly cited researcher by Clarivate in 2020-2022.
\end{IEEEbiography}

\begin{IEEEbiography}[{\includegraphics[width=1in,height=1.25in,clip,keepaspectratio]{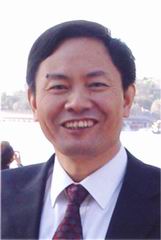}}]{Yaoxue Zhang [M’17, SM’18]}
 received the B.Sc. degree from the Northwest Institute of Telecommunication Engineering, Xi’an, China, in 1982, and the Ph.D. degree in computer networking from Tohoku University, Sendai, Japan, in 1989. He is currently a professor with the Department of Computer Science and Technology, Tsinghua University, China. His research interests include computer networking, operating systems, and transparent computing. He has published more than 200 papers on peer-reviewed IEEE/ACM journals and conferences. He is the editor-in-chief of Chinese Journal of Electronics and a fellow of the Chinese Academy of Engineering. 
\end{IEEEbiography}

\end{document}